\documentclass[10pt,twocolumn,letterpaper]{article}

\usepackage[pagenumbers]{cvpr2024_conference}      
\usepackage{tcolorbox}
\usepackage{multirow}
\usepackage{colortbl} 
\usepackage{amsmath}
\usepackage{amssymb}
\usepackage{booktabs}
\usepackage{tabularray}
\usepackage{graphicx, subcaption}

\definecolor{mygray}{gray}{0.6}
\definecolor{mygray-bg}{gray}{0.9}
\makeatletter\renewcommand\paragraph{\@startsection{paragraph}{4}{\z@}
	{.25em \@plus1ex \@minus.2ex}{-.5em}{\normalfont\normalsize\bfseries}}\makeatother
\definecolor{cvprblue}{rgb}{0.21,0.49,0.74}
\usepackage[pagebackref,breaklinks,colorlinks,citecolor=cvprblue]{hyperref}

\title{\textit{Audio-Visual LLM} for Video Understanding}

\author{%
  Fangxun Shu$^{1}$\thanks{Equal contribution.} \quad \quad \quad
  Lei Zhang$^{1,2*}$ \quad \quad \quad 
  Hao Jiang$^{1}$\thanks{Corresponding author.} \quad \quad \quad
  Cihang Xie$^{3\dag}$ \vspace{.3em}\\
    $^{1}$Alibaba Group ~~ $^{2}$Zhejiang University ~~ $^{3}$University of California, Santa Cruz \vspace{-.3em} \\
}

\begin{document}
\maketitle
\begin{abstract}

This paper presents Audio-Visual LLM, a Multimodal Large Language Model that takes both visual and auditory inputs for holistic video understanding. A key design is the modality-augmented training,
which involves the integration of modality-specific tokens engineered to activate the appropriate visual and/or auditory encoder selectively. This mechanism is pivotal in enabling end-to-end joint training with video data at different modalities, including visual-only, audio-only, and audio-visual formats. 
Moreover, we introduce a high-quality video instruction dataset, derived from GPT-4. This dataset allows Audio-Visual LLM to adeptly process a variety of task-oriented video instructions, ranging from multi-turn conversations and audio-visual narratives to complex reasoning tasks. 

Extensive experiments demonstrate that Audio-Visual LLM impressively achieves strong zero-shot results across a range of video understanding tasks. For example, Audio-Visual LLM achieves an accuracy of 53.7\% on MSRVTT-QA, outperforming non-LLM-based InterVideo by 6.6\% and LLM-based Valley by 4.4\%, respectively. Additionally, our Audio-Visual LLM also achieves competitive performance on audio tasks (\eg, AudioCaps).

\end{abstract}
\section{Introduction}
\label{sec:intro}

\begin{figure}
    \centering
    \includegraphics[width=1.0\linewidth]{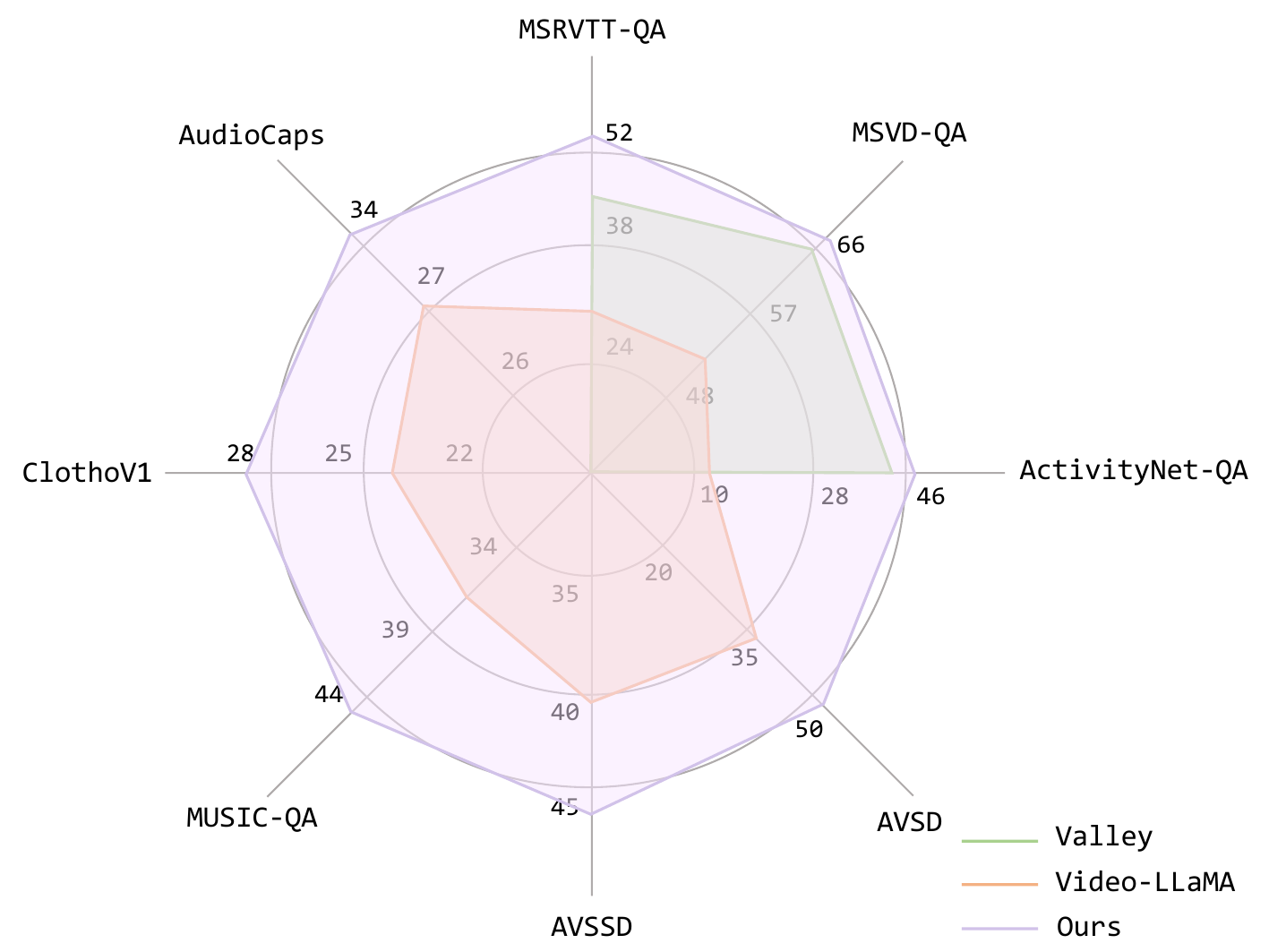}
    \vspace{-2em}
    \caption{Audio-Visual LLM achieves advantageous performance on various video understanding tasks and consistently outperforms existing methods.}
    \vspace{-1em}
    \label{fig:enter-label}
\end{figure}

Videos are inherently multimodal, encapsulating both auditory and visual information. This multimodality is not just an inherent characteristic of videos but also a fundamental aspect of how humans perceive and interact with visual media. For instance, in a cinematic context, the simultaneous engagement with both visual imagery and auditory cues significantly enriches the viewing experience, enhancing both comprehension and enjoyment. Drawing inspiration from this intrinsic human experience, empowering multimodal models~\cite{clipbert, hero, UniVL, mac, alpro} to concurrently understand visual and audio leads to substantial improvements in video understanding.

Large language models~\cite{gpt-3, palm, llama, glm} have shown significant abilities in intent understanding and instruction following. They can interact with human intents and provide appropriate feedback based on the given instructions. Research works~\cite{flamingo, instructblip, blip2} further extend to LLMs with visual perception abilities, designing alignment modules and curating instruction-following datasets, particularly in image understanding~\cite{minigpt,llava, otter,qwen-vl} and video understanding~\cite{valley, video-chatgpt, videochat}. 
However, these early efforts mostly focus on visual content, thereby underutilizing the rich auditory data present in videos.

To bridge this gap, recent works~\cite{videollama, macawllm, pandagpt} have begun to incorporate visual and audio components in enhancing video understanding. However, these models still exhibit significant limitations in the joint processing of audio-visual data. Video-LLaMA's~\cite{videollama} approach to audio signal representation and alignment is potentially limited, as it primarily relies on the capability of the pre-trained ImageBind model \cite{imagebind}.
Similarly, MacawLLM~\cite{macawllm} adopts visual and audio signals extracted from videos of different sources, which may induce bias and instability in training. 
These observations highlight a substantial opportunity for improvement in aligning audio-visual modalities, both in terms of model architecture and dataset development.


To this end, we hereby introduce a multimodal LLM framework that synergistically aligns visual and audio signals for holistic video understanding. This framework includes two key contributions. First, we implement a modality augmentation approach during the training of the Audio-Visual LLM. This technique facilitates a comprehensive exploration of the interplay between visual and audio signals in videos. As the training batch consists of visual, audio, and audio-visual samples, imitating the component of video, we incorporate modality-specific tokens to activate the appropriate visual and/or audio encoders selectively. This mechanism is pivotal in enabling end-to-end joint training to flexibly fuse different modalities in video data. 
Secondly, we present a pipeline to curate visual/audio-text pairs into the appropriate and diverse instruction-following format, using GPT-4~\cite{gpt4}. To comprehensively enhance the training of Audio-Visual LLM, we meticulously curate general captions including long descriptions, for the pre-training stage and fine-grained instructions including multi-turn conversations and complex reasoning, for the instruction fine-tuning stage.

Extensive experiments demonstrate that our method achieves advantageous performance on various video understanding tasks. For instance, we achieve an accuracy of 53.7\% on MSRVTT-QA and 47.2\% on ActivityNet-QA, outperforming both LLM-based  (\eg, Video-LLaMA~\cite{videollama} and Valley~\cite{valley}) and non-LLM-based (\eg, VideoCoCa~\cite{yan2022video} and InterVideo~\cite{wang2022internvideo}) comparison works. We also show comparable performance on audio tasks (\eg, AudioCaps~\cite{audiocaps}), demonstrating the substantial potential of our method in audio. 
\vspace{-1mm}\section{Related Work}
\label{sec:related}

\begin{figure*}[t]
  \centering
  \scalebox{0.95}{
   \includegraphics[width=1\linewidth]{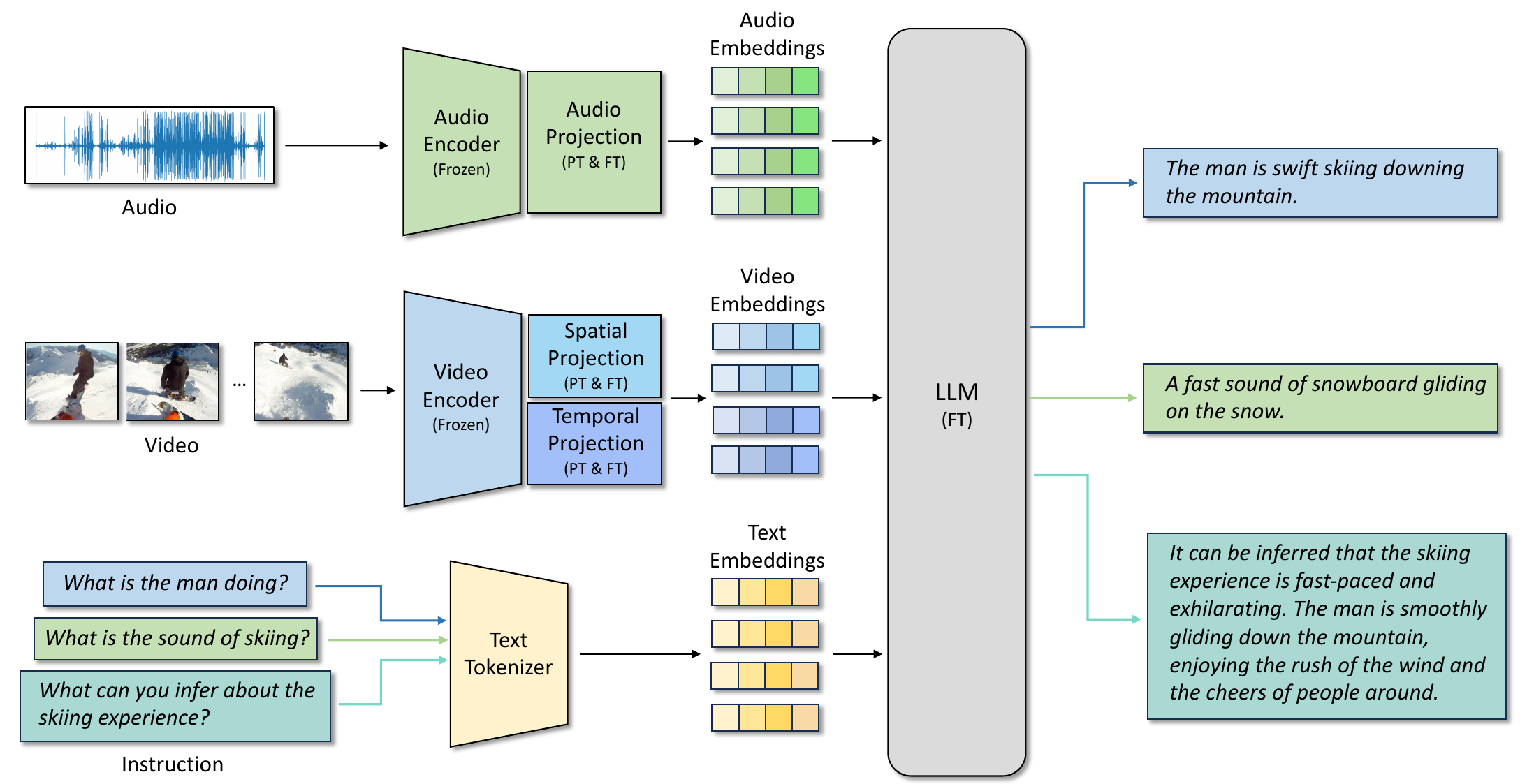}
   }
   \vspace{-.7em}
   \caption{The framework of our method consists of three components: multimodal encoders, linear projections, and LLM. We pre-train the projections in the pre-training stage and fine-tune both the projections and LLM in the sft stage. Both stages freeze the multimodal encoders.}
   \vspace{-2mm}
   \label{fig:method}
\end{figure*}

\paragraph{Multimodal Video Understanding.}
Video, as a multimodal medium, contains rich visual, language, and audio information. Prior works focus on the alignment and fusion of modalities within videos through the pre-training on the large-scale dataset~\cite{howto100m,acav100m,Frozen,yt180m,videocc}, including domains of video~\cite{videomae, st_mae, bevt, omnimae, vivit, timesformer}, video-language~\cite{videobert,mac,videoclip,hero,clipbert}, and video-audio-language~\cite{vatt,video-audio-vit,avmae,cvmae,polyvit,chen2023valor}. Recently, the significant advancements in generative large language models~\cite{gpt1,gpt2,gpt-3} provide inspiration for multimodal video generative pre-training. VideoMAE~\cite{videomae} extends the asymmetric encoder-decoder architecture in MAE~\cite{mae} and predicts the pixels for masked patches. BEVT~\cite{bevt} leverages a pre-trained VQ-VAE~\cite{vqvae} as the tokenizer to generate discrete visual tokens and predict the tokens for masked patches. VIOLET~\cite{violet} and MILES~\cite{miles} extend this paradigm to the video-language domain. With the emergence of large language models, we leverage the instruction-following capability of LLMs to comprehensively enhance video understanding.

\paragraph{Multimodal Large Language Models.}
Drawing inspiration from the powerful instruction-following~\cite{instructgpt} capability of large language models (LLMs)~\cite{gpt1,gpt2,gpt-3}, recent studies extend LLMs to understand multimodal content. A series of works enable LLMs to understand visual or audio content~\cite{llava,minigpt,otter,video-chatgpt,videochat,instructblip}. These works focus on designing projection modules to align multimodal data with LLMs, as well as constructing multimodal instruction datasets to enhance the LLMs' ability to follow instructions for multimodal data. LLaVA~\cite{llava} connects images into LLMs with a linear projector, using the image instruction datasets curated via GPT-4~\cite{gpt4}. Valley~\cite{valley} employs a simple pooling operation to unify images and videos into LLMs. AudioGPT~\cite{audiogpt} leverages various audio foundation models to process audio data, where LLMs are regarded as the general-purpose interface. Recently, some studies enable LLMs to jointly understand visual and audio content within videos~\cite{videollama, macawllm, pandagpt}. Video-LLaMA~\cite{videollama} directly leverages pre-trained ImageBind~\cite{imagebind} as a universal embedding space to align visual and audio modalities. MacawLLM~\cite{macawllm} designs a learnable alignment module and uses visual and audio data as training samples. However, the visual and audio signals adopted are extracted from videos of different sources, which induces bias and instability in training. However, the alignment between visual and audio modality remains under-explored. We propose a modality-augmented training strategy and meticulously curate visual-audio instruction datasets to thoroughly explore the visual and audio alignment.
\section{Methods}
\label{sec:method}

\begin{figure*}[h]
    \centering
    \scalebox{0.9}{
    \includegraphics[height=0.5\textheight]{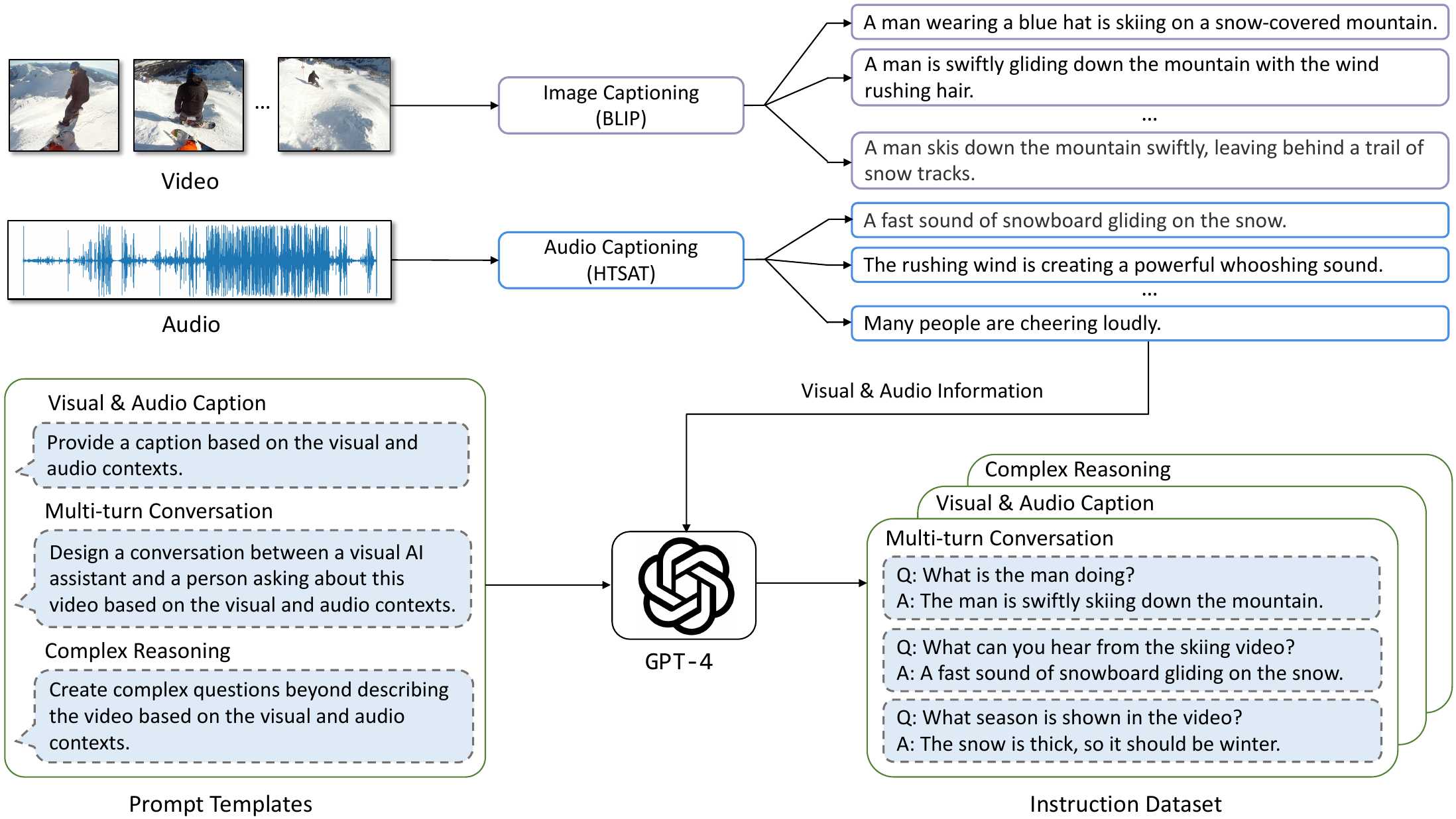}
    }
    \vspace{-2em}
    \caption{The framework of the GPT-assisted automated data curation. We utilize pre-trained visual and audio models to generate rich visual and audio contexts. We then design various prompt templates to generate high-quality instruction datasets, including visual \& audio captions, multi-turn conversation, and complex reasoning.
    }~\label{data_construction}
        \vspace{-2mm}
\label{fig:data_generation}
\end{figure*}

In this section, we first introduce the overall architecture of our Audio-Visual LLM as shown in~\cref{fig:method}. Secondly, we propose a modality-augmented training strategy to improve the visual and audio alignment. Lastly, we curate a high-quality instruction dataset in the context of visual-only, audio-only, and joint audio-visual as shown in~\cref{fig:data_generation}, facilitating the modality-augmented training.

\vspace{-1mm}\subsection{Model Architecture}
\label{sec:model-design}
Our model consists of three components: the multimodal encoders, the linear projectors, and the large language model, as illustrated in~\cref{fig:method}.


\paragraph{Multimodal Encoders.} Given a video, we first decompose it into individual video frames $F \in \mathbb{R}^{T \times H \times W \times 3}$ and audio segments $A \in \mathbb{R}^{K \times M}$, where $T$ represents the frame number, and $K$ represents the segment number. For video frames, we divide each frame into $N$ non-overlapping patches in the spatial dimension and generate spatio-temporal patches $V \in \mathbb{R}^{T \times N \times P \times P \times C}$, where $P$ represents the patch size. We utilize CLIP~\cite{clip} to independently encode frame-level embeddings $E_{v} \in \mathbb{R}^{T \times N \times D}$. Instead of directly feeding the embeddings into LLMs, we propose a more flexible solution, which can process long videos. Given $E_{v}$ with a sequence length of $T \times N$, we concatenate the \verb+[CLS]+ token of each frame embedding to form the temporal tokens $E_{t} \in \mathbb{R}^{T \times D_{t}}$, where $D_{t}$ represents the embedding dimension of the temporal tokens. Simultaneously, we pool the patch embeddings along the temporal dimension to form the spatial tokens $E_{s} \in \mathbb{R}^{N \times D_{s}}$, where $D_{s}$ represents the embedding dimension of the spatial tokens. The final sequence length of temporal tokens $E_{t}$  and spatial tokens $E_{s}$  that are fed into LLMs is reduced to $T+N$, as shown below:

\begin{equation}
E_{t} = \{v_{cls}^{1}, v_{cls}^{2},..., v_{cls}^{T}\}, 
\label{eq:temporal-tokens}
\end{equation}
\begin{equation}
E_{s} = \{\bar{v}^{1}, \bar{v}^{2}, \bar{v}^{3}, ..., \bar{v}^{N}\}. 
\label{eq:spatial-tokens}
\end{equation}

For audio segments, we utilize CLAP~\cite{clap} to extract the last hidden state as the audio embedding, which encapsulates the information regarding the semantic modeling of the audio segments. Subsequently, we flatten this hidden state into $E_{a} \in \mathbb{R}^{K \times D_{a}}$, where $D_{a}$ represents the embedding dimension for each audio segment:

\begin{equation}
E_{a} = \{{a}^{1}, {a}^{2}, {a}^{3}, ..., {a}^{K}\}.  
\label{eq:auditory-tokens}
\end{equation}

\paragraph{Linear Projectors.} We use linear layers as the projectors to transform the temporal tokens $E_{t} \in \mathbb{R}^{T \times D_{t}}$, spatial tokens $E_{s} \in \mathbb{R}^{N \times D_{s}}$, and auditory tokens $E_{a} \in \mathbb{R}^{K \times D_{a}}$ to match the embedding dimension $D_{l}$ of LLMs. The transform is denoted below, where $W$ and $b$ are weight matrices and bias vectors, respectively.

\begin{equation}
\begin{aligned}
H_{t} &= W_{t}E_{t} + b_{t}, \\
H_{s} &= W_{s}E_{s} + b_{s}, \\
H_{a} &= W_{a}E_{a} + b_{a}.
\end{aligned}
\end{equation}

The final video tokens to the LLM are denoted as $H_{v} \in \mathbb{R}^{(T+N+K) \times D_{l}}$, where the first $T$ tokens represent temporal tokens $H_{t}$, the next $N$ tokens represent spatial tokens $H_{s}$, and the last $K$ tokens represent auditory tokens $H_{a}$.


\paragraph{Large Language Model.} We build upon the open-source Vicuna~\cite{vicuna} as the foundation for our large language model, which is fine-tuned on LLaMA~\cite{llama} with approximately 70,000 user-shared dialogues collected from ShareGPT. We combine Instruction tokens $I$ with the video tokens $H_{v}$ and feed them into Vicuna. This combination allows the model to generate a coherent textual response $R$ that aligns with the given video, as illustrated below:
\begin{equation}
R = \mathrm{LLM}(H_{v}, I).
\label{eq:llm-interface}
\end{equation}

 \begin{figure}[h]
    \centering
    \scalebox{0.98}{
    \includegraphics[width=1.0\linewidth]{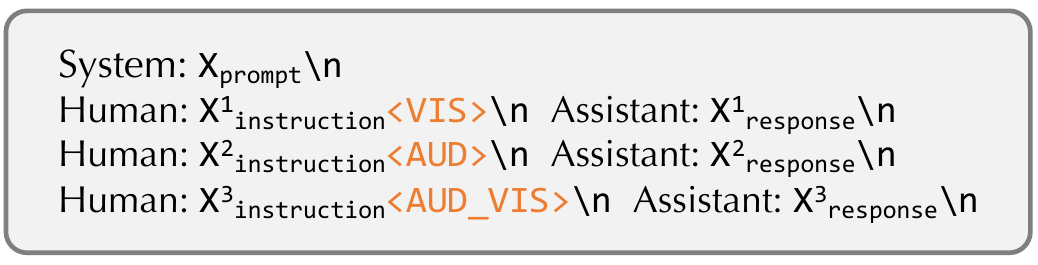}
    }
    \vspace{-2em}
    \caption{Multi-turn video instructions. $\texttt{X}_{\texttt{prompt}}$ is the prompt template. \texttt{<VIS>}, \texttt{<AUD>}, and \texttt{<AUD\_VIS>} are the modality-specific tokens for visual, audio, and audio-visual samples in videos. During the training stage, we activate the encoder and projector to extract embeddings according to the modality-specific token.
    }
    \label{fig:instruction_example}
    \vspace{-5mm}
\end{figure}

\subsection{Modality-Augmented Training}
\label{sec:training-procedure}

Video inherently contains both visual signals, audio signals, and audio-visual signals. We hereby propose a novel training paradigm, termed Modality-Augmented Training (MAT), to jointly train three modal types of samples (\ie, visual-only, audio-only, and audio-visual joint samples) within a single batch. This paradigm allows our model to simultaneously consider multiple perspectives of video, enabling a more comprehensive understanding of its content. 
Specifically, we incorporate a modality-specific token $ \texttt{<MOD>} \in \left \{ \texttt{<VIS>}, \texttt{<AUD>}, \texttt{<AUD\_VIS>} \right \} $ to denote the visual and audio content in the prompt, as shown in ~\cref{fig:instruction_example}. This token will be replaced with the modal sample, and the appropriate encoder and projector will be activated to obtain the embeddings. For instance, given a vision-only instruction data, $\texttt{<MOD>}$ is set as $\texttt{<VIS>}$. When the model parses the input prompt, $\texttt{<VIS>}$ will be replaced with the visual sample in the video, and the model will activate the visual encoder and projector to compute the visual embeddings. We set the audio embedding as None for the input so that the audio encoder and projector will not be updated during the backward propagation.

To comprehensively enhance the training effectiveness, we adopt a two-stage training approach. In the first stage, we meticulously curate general captions, including lengthy descriptions, for pre-training. In the second stage, we prepare detailed instructions that cover multi-turn conversations and complex reasoning for instruction fine-tuning.

\subsection{GPT-Assisted Data Curation}
\label{sec:data-constrcution}

In this section, we introduce a GPT-assisted data curation approach as illustrated in~\cref{fig:data_generation}. We utilize pre-trained models~\cite{blip,wavcaps} to generate audio-visual textural contexts and design prompt templates to guide GPT-4~\cite{gpt4} in generating visual, audio, and audio-visual instruction. Detailed explanations of each component are shown below.

\paragraph{Visual \& Audio Contexts.} We leverage datasets, the ACAV100M~\cite{acav100m} dataset and the VGGSound~\cite{vggsound} dataset to curate video instructions from different modalities. The visual signals and audio signals are both available within one single video, which avoids the bias from different videos. Additionally, we augment the dataset by incorporating the WebVid2M~\cite{Frozen} dataset for visual-only instructions and the WavCaps~\cite{wavcaps} dataset for audio-only instructions. For visual context, we use a pre-trained BLIP~\cite{blip} model to extract frame captions, resulting in a set of frame captions $V_{c}=\{v_{c}^{1}, v_{c}^{2}, ..., v_{c}^{T}\}$. For audio context, we use a pre-trained HTSAT~\cite{wavcaps} model to extract audio segment captions, resulting in a set of audio captions $A_{c}=\{a_{c}^{1}, a_{c}^{2}, ..., a_{c}^{K}\}$.



\paragraph{Prompt Templates.} We design intricate prompt templates to guide instruction generation. The prompt template is divided into four sections: ROLE, REQUIREMENT, EXAMPLE, and CONTEXT. The ROLE section establishes GPT-4 as a video AI assistant. The REQUIREMENT section states our detailed conditions for different tasks. The EXAMPLE section presents a specific question-answer pair to demonstrate the desired style and format of the generated responses. The CONTEXT section accepts the textural contexts as the inputs. We thereby input visual textual contexts $V_{c}$ and audio textual contexts $A_{c}$ into prompts to GPT-4~\cite{gpt4} to generate various high-quality datasets. The details and examples of prompt templates can be found in the supplementary material.

\paragraph{Instruction Datasets.} By setting the REQUIREMENT section in the prompt template, our instruction dataset consists of diverse tasks as follows: 
\begin{itemize}
    \item \textbf{Audio-visual caption.} It describes spatial and temporal relationships of objects in video, and auditory cues in audio, which empowers the model to capture general visual and audio knowledge.
    \item \textbf{Multi-turn Conversation.} This involves iterative audio-visual interactions between the person and the assistant and consists of multi-level and multi-granularity details, which empowers the model with fine-grained understanding and instruction-following.
    \item \textbf{Complex Reasoning.} This surpasses simple factual responses and engages in higher-level reasoning, such as sentimental inferring, which empowers the model with the capability of reasoning.
\end{itemize}

\section{Experiments}
\label{sec:experiments}

In this section, we first introduce the experimental setups and implementation details. We then describe the downstream tasks that our method is evaluated on and report strong results and ablation studies.

\subsection{Experimental Setup}

\paragraph{Model Settings.} We build the visual encoder with ViT-L/14~\cite{vit}, which is a transformer-based model comprised of 24 layers of blocks, with a patch size of 14. We initialize it from the pre-trained CLIP~\cite{clip} version via contrastive learning. Similarly, we build the audio encoder with HTSAT~\cite{HTSAT}, which is also a transformer-based model with 4 groups of swin transformer blocks~\cite{swin}. We initialize it from the pre-trained CLAP~\cite{clap} via contrastive learning. We use fully connected layers as linear projectors to convert spatial, temporal, and audio tokens to the LLM with a dimension of 4096. We build our LLM with Vicuna-7B~\cite{vicuna}, which is an effective chat version that has been fine-tuned on LLaMA~\cite{llama}.

\paragraph{Training Datasets.}  We curate 260k video instruction data including 100k captions and 160k instructions. We also use 650k image data from LLaVA, and 770k video data from Valley to enhance the shared spatio-temporal perception and reasoning capacity.

\paragraph{Implementation Details.} We resize videos to the resolution of 224$\times$224 and uniformly sample 32 frames during training. We evenly divide audio signals into 4 segments and sample each segment at a sampling rate of 48Khz. We treat images as 1-frame videos so that we can jointly train images and videos in a unified manner. We adopt FlashAttention~\cite{flashattention} and ZeRO~\cite{zero} for efficient training. We use the AdamW optimizer~\cite{adamw} with $\beta=(0.9, 0.98)$. A cosine annealing learning rate schedule is applied with a warmup ratio of 0.03. The training is conducted on 8$\times$A100 GPUs with 80GB GPU memory. During the pre-training stage, we freeze the encoders and LLM and only train the linear projectors. The learning rate is set to 2$\mathrm{e}$-3. We use a total batch size of 256 and set the training epoch of 3, taking approximately 16 GPU hours. During the instruction fine-tuning stage, we only freeze the encoders and jointly train the linear projectors and LLM. The learning rate is set to 2$\mathrm{e}$-5. We use a total batch size of 128 and set the training epoch of 1, taking approximately 10 GPU hours. 

\subsection{Results}
\label{sec:results}

\paragraph{Downstream Tasks and Datasets}
We explore our method on 6 video understanding tasks, covering video question-answering (QA) and audio-visual question-answering (QA). We also evaluate our method on 2 audio captioning (AC) tasks to demonstrate the generalization of audio understanding. Unless stated otherwise, we report top-1 accuracy~\cite{video-chatgpt} on QA tasks and CIDEr~\cite{cider} \& SPIDEr~\cite{spider} on AC tasks, respectively. 

\begin{itemize}
    \item \textbf{Video QA.} We evaluate video QA tasks in an open-ended manner, covering 3 public datasets: MSVD-QA~\cite{msvd}, MSRVTT-QA~\cite{msrvtt}, and ActivityNet-QA~\cite{activitynet}.
    \item \textbf{Audio-visual QA.} We evaluate audio-visual QA tasks in an open-ended manner, covering 3 public datasets: AVSD~\cite{avsd}, AVSSD~\cite{vggsound}, and MUSIC-AVQA~\cite{musciqa}.
    \item \textbf{Audio Captioning.} We evaluate AC tasks on 2 public datasets: ClothoV1~\cite{clotho} and AudioCaps~\cite{audiocaps}.
\end{itemize}

\begin{table*}[]
\centering
\scalebox{0.85}{
\begin{tabular}{lllllll}
\toprule
\textbf{Method}    &  \textbf{\# PT Datasets}   & \textbf{\# PT Pairs} & \textbf{Modality} & \textbf{MSRVTT-QA}  & \textbf{MSVD-QA}   & \textbf{ActivityNet-QA}  \\
\midrule
\multicolumn{1}{l}{\textbf{\textit{w/o LLM, PT+FT}}} \\
QueST~\cite{quest}   & {--}     & {--} &  {V}     & {34.6}       & {34.6}    &    --     \\
ClipBERT~\cite{clipbert}  & {VG, COCO}  & {5.6M} & {V}    & {37.4}       & --       &  --     \\
JustAsk~\cite{yang2021just}   & {HTVQA}   & 69M  &V   & 41.5          & 46.3                & 38.9  \\
GIT~\cite{wang2022git}     & {CC, VG, SBU, COCO}      & 20M     & V       & 42.7       & 55.1        & --    \\
MERLOT~\cite{zellers2021merlot}  &  {YT}    & 180M &V     & 43.1    & --         & 41.4     \\
Singularity~\cite{lei2022revealing} & {CC, WV}   & 17M &V      & 43.5       &   --           & 43.1  \\
Clover~\cite{huang2022clover}   &  {CC, WV}  & {5.5M} & {V}    & {43.9}      & 51.9      &    --      \\
VIOLET~\cite{violet}  &  {CC, WV, YT} & {185.5M} & {V}    & {44.5}     & 54.7       &   --    \\
LAVENDER~\cite{li2022lavender}   & {CC, WV, VG, SBU, COCO}   & 30M &V      & 45.0           & 56.6          &  --    \\

VideoCoCa~\cite{yan2022video}  &  {HT, VCC}  & 140M    & V       & 46.0           & 56.9  & --  \\
VALOR~\cite{chen2023valor} &  {CC, WV, VAL}   & 6.5M    & V, A    & 46.7   &  56.4      &   44.8   \\
All-in-one~\cite{wang2022all}  &   {WV, HT}  & 122.5M &V   & 46.8    & 48.3         &  40.0   \\
FrozenBiLM~\cite{frozenbilm}   & {WV}        & 10M     & V       & 47.0     & 54.8    & 43.2    \\

InterVideo~\cite{wang2022internvideo} &  {WV, HT, LAION}  & 147M & V & 47.1    & 55.5 & --            \\

\midrule
\multicolumn{1}{l}{\textbf{\textit{w LLM, PT+SFT}}}  \\
{LLaMA-Adapter~\cite{llama-adapter}} &  {CC, COCO}   & {0.6M} & {V}    & {43.8}  & {54.9}  & {34.2}  \\
{VideoChat~\cite{videochat}} &  {CC, WV, VG, SBU, COCO}  & {25M} & {V}    & {45.0}    & {56.3}  & {26.5}  \\
{Video-ChatGPT~\cite{video-chatgpt}} &  {CC, AN, COCO,} & {0.85M} & {V}  & {49.3} & {64.9}  & {35.2}  \\
{Valley~\cite{valley}} &  {CC, WV, COCO}  & {1.5M} & {V} & {45.7}  & {65.4}& {42.9} \\
{Video-LLaMA~\cite{videollama}} & {CC, WV, COCO} & {2.8M} & {V, A} & {29.6}  & {51.6} & {12.4}  \\
{PandaGPT~\cite{pandagpt}} &  {CC, SBU, COCO, LAION} & {128M} & {V, A} & {23.7}  & {46.7} & {11.2}  \\
{MacawLLM~\cite{macawllm}} &  {COCO, AVSD}  & {0.3M} & {V, A} & {25.5}  & {42.1} & {14.5} \\
\rowcolor{mygray-bg}{Ours} &  {CC, WV, VS, WC, ACAV, COCO} & {1.6M} & {V, A}  & {\textbf{53.7\textcolor{red}{$_{+4.4}$}}}  & {\textbf{67.3\textcolor{red}{$_{+1.9}$}}} & {\textbf{47.2\textcolor{red}{$_{+2.4}$}}} \\

\bottomrule
\end{tabular}}
\vspace{-.7em}
\caption{Comparison with state-of-the-art methods on 3 open-ended video QA benchmarks. \textbf{\# PT Pairs}: number of pairs for pre-training. \textbf{\# Modality}: modalities that the model can handle, where \textbf{V} for video and \textbf{A }for audio. CC, WV, HT, VS, WC, ACAV, VAL, VCC are short for C C14M~\cite{sharma2018conceptual}, WebVid~\cite{Frozen}, HowTo100M~\cite{howto100m}, VGGSound~\cite{vggsound}, WavCaps~\cite{wavcaps}, ACAV100M~\cite{acav100m}, VALOR1M~\cite{chen2023valor}, and VideoCC3M~\cite{videocc}, respectively.}
\label{tab:video-qa}
\end{table*}

\begin{table}[]
\centering
\scalebox{0.9}{
\begin{tabular}{l|ccc}
\toprule
\textbf{Method}    & \textbf{AVSD}   & \textbf{AVSSD}   & \textbf{MUSIC-QA} \\
\midrule
{Video-LLaMA~\cite{videollama}}  & {36.7}  & {40.8} & {36.6}  \\
{PandaGPT~\cite{pandagpt}}  & {26.1}  & {32.7}  & {33.7}  \\
{McawLLM~\cite{macawllm}}   & {34.3}  & {36.1}  & {31.8}  \\
\rowcolor{mygray-bg}{Ours (video \& audio)}  & 52.6 & 47.6 & 45.2  \\
\bottomrule
\end{tabular}}
\vspace{-.7em}
\caption{Comparison with existing LLM-based methods on 3 open-ended audio-visual QA benchmarks.}
\label{tab:avqa_results}
\end{table}

\begin{table}[]
    \centering
    \scalebox{0.83}{ 
    \begin{tabular}{l|cccc}
    \toprule
    \multirow{2}{*}{\textbf{Method}} & \multicolumn{2}{c}{\textbf{ClothoV1}} & \multicolumn{2}{c}{\textbf{AudioCaps}}  \\
    {} & {\textbf{CIDEr}} & {\textbf{SPIDEr}} & {\textbf{CIDEr}} & {\textbf{SPIDEr}} \\
    \midrule
    {Video-LLaMA~\cite{videollama}} & {24.5} & {15.2} & {27.6} & {17.8} \\
    {PandaGPT~\cite{pandagpt}} & {21.5} & {13.4}  & {29.1} & {19.6} \\
    {McawLLM~\cite{macawllm}} & {26.1} & {17.7} & {33.3} & {21.4} \\
    \rowcolor{mygray-bg}{Ours (audio-only) } & 29.6  & 19.7 & 35.4 & 24.1\\
    \bottomrule
    \end{tabular}
    }
    \vspace{-.7em}
    \caption{Comparison with existing LLM-based methods on 2 audio captioning benchmarks.}
    \label{tab:audio_caption}
    \vspace{-4mm}
\end{table}

\paragraph{Video QA.} \cref{tab:video-qa} shows the video QA results on the MSRVTT, MSVD, and ActivityNet datasets, where the average video durations are 10s, 15s, and 180s, respectively. This allows for a comprehensive evaluation of short-term and long-term spatio-temporal understanding. The results demonstrate that our method surpasses both prior non-LLM-based works and LLM-based works across all the datasets by a large margin. 

Compared to the prior non-LLM-based works, we observe that our method brings a +6.6\% accuracy on MSRVTT-QA, a +10.4\% accuracy on MSVD-QA, and a +2.4\% accuracy on ActivityNet-QA. From the significant improvements, we can find that our method, which fine-tunes LLM on a small amount of instruction data ($\sim$1M), efficiently achieves better performance than the non-LLM-based works that pre-training with the large-scale dataset ($\sim$100M). We analyze that our method leverages the CLIP-L/14, which has been pre-trained on a massive dataset of 400M image-text pairs, as the visual encoder. The visual embeddings obtained from this encoder have been effectively aligned with the textual embedding space. Therefore, we can align videos with LLM using a relatively small amount of instruction dataset, harnessing the powerful instruction-following capability of LLMs for effective video comprehension. 

Compared to the prior LLM-based works that support video-only input and audio-visual input, our method consistently brings a +4.4\% accuracy on MSRVTT-QA, a +1.9\% accuracy on MSVD-QA, and a +2.4\% accuracy on ActivityNet-QA. From the significant improvements, we can find that the proposed modality-augmented training mechanism, which jointly optimizes diverse modality samples in the same video can significantly enhance video alignment with LLMs compared to works (\eg, Valley and Video-ChatGPT) that focus on visual-only samples. Moreover, we find that the high-quality video instruction dataset plays a crucial role. For example, PandaGPT only uses the image instruction dataset to optimize the ImageBind, while Video-LLaMA leverages the video caption dataset during pre-training, yet still relies on the image instruction dataset during instruction fine-tuning, resulting in the loss of temporal information and sub-optimal outcomes.



\paragraph{Audio-Visual QA.} \cref{tab:avqa_results} shows the audio-visual QA results on the AVSD, AVSSD, and MUSIC-QA. We find that PandaGPT and Video-LLaMA use ImageBind as the audio encoder to optimize visual and audio alignment with LLMs in cases where there is a lack of audio instruction dataset, inherently losing audio alignment. MacawLLM uses a stronger Whisper as the audio encoder while using video and audio instruction data from different videos to optimize visual and audio branches separately, which is insufficient in cross-modality alignment within the same video. Compared to the prior LLM-based works, our method performs modality-augmented training on the dataset with audio-visual instructions which has an advancing cross-modality understanding within videos. The results show that our method surpasses them by a large margin with a +15.9\% accuracy on AVSD, a +6.8\% accuracy on AVSSD, and a +8.6\% accuracy on MUSIC-QA. 

\paragraph{Audio Captioning.} While our method primarily focuses on video understanding, our method also supports audio-only input during the inference stage, enabling us to handle various audio tasks. We conduct experiments on audio captioning tasks, including ClothoV1 and AudioCaps. Prior works such as Video-LLaMA and PandaGPT often struggle with audio alignment, mainly due to the lack of corresponding instructions. We leverage the curated audio instructions provided by WavCaps, leading to comparable audio instruction-following capability. The experimental results in~\cref{tab:audio_caption} show that our method performs consistent advantages over prior works by a +3.5\% CIDEr on ClothoV1 and a +2.1\% CIDEr on AudioCaps, which demonstrates the potential of our method in the audio domain.

\subsection{Ablation Studies}
\label{sec:ablation}
In this section, we investigate the effects of various design choices including modality, architecture, and data. By default, we use ViT/L-14 as a visual encoder, HTSAT as an audio encoder, and Vicuna as the LLM, respectively. The video is sampled with 32 frames and 4 audio segments during training and inference.

\paragraph{Training Strategy.} 

\begin{figure}[t]
  \centering
  \subcaptionbox{MSRVTT-QA}{
    \includegraphics[width=.46\linewidth]{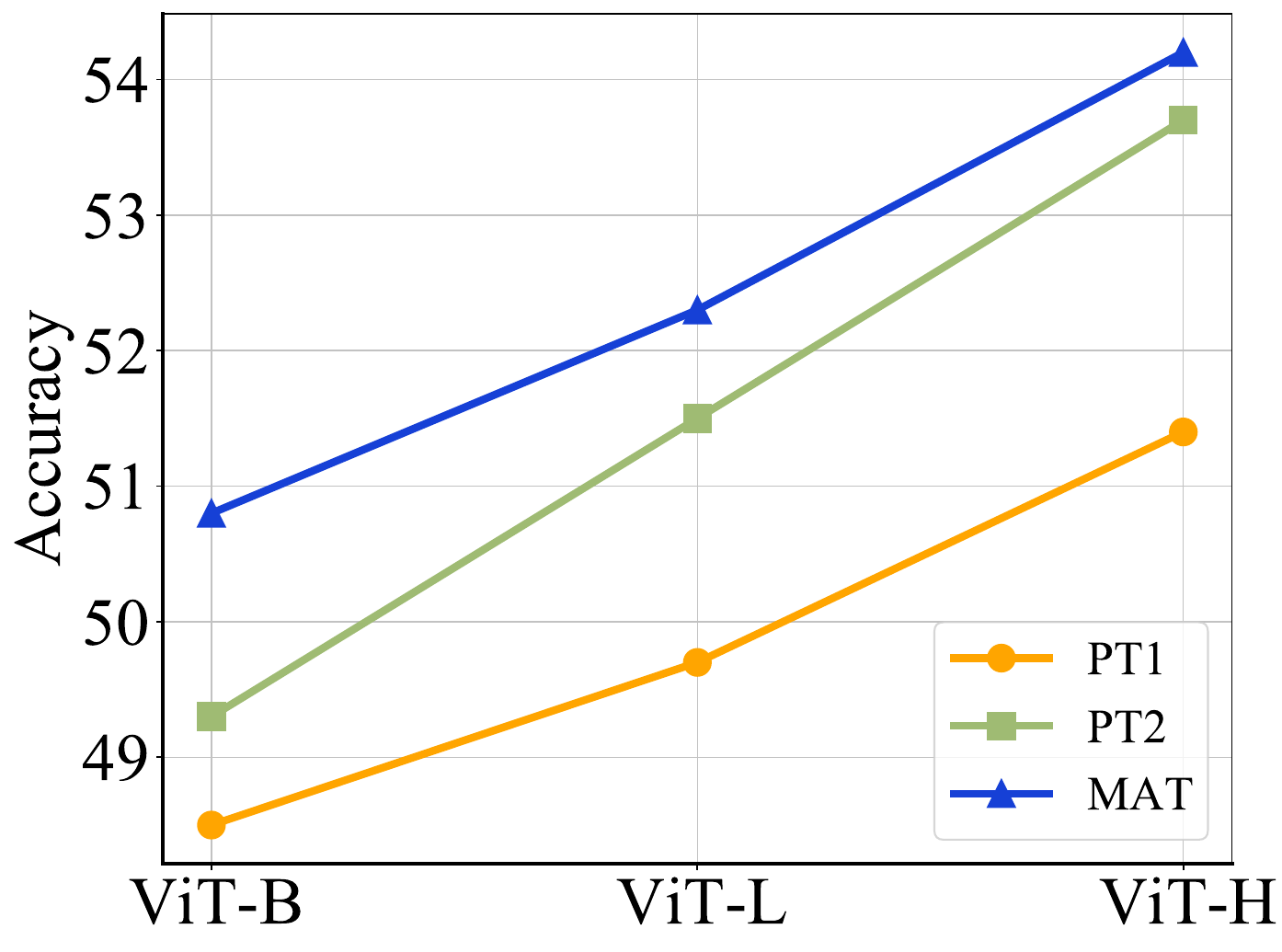}
  }
  \subcaptionbox{AVSD}{
    \includegraphics[width=.46\linewidth]{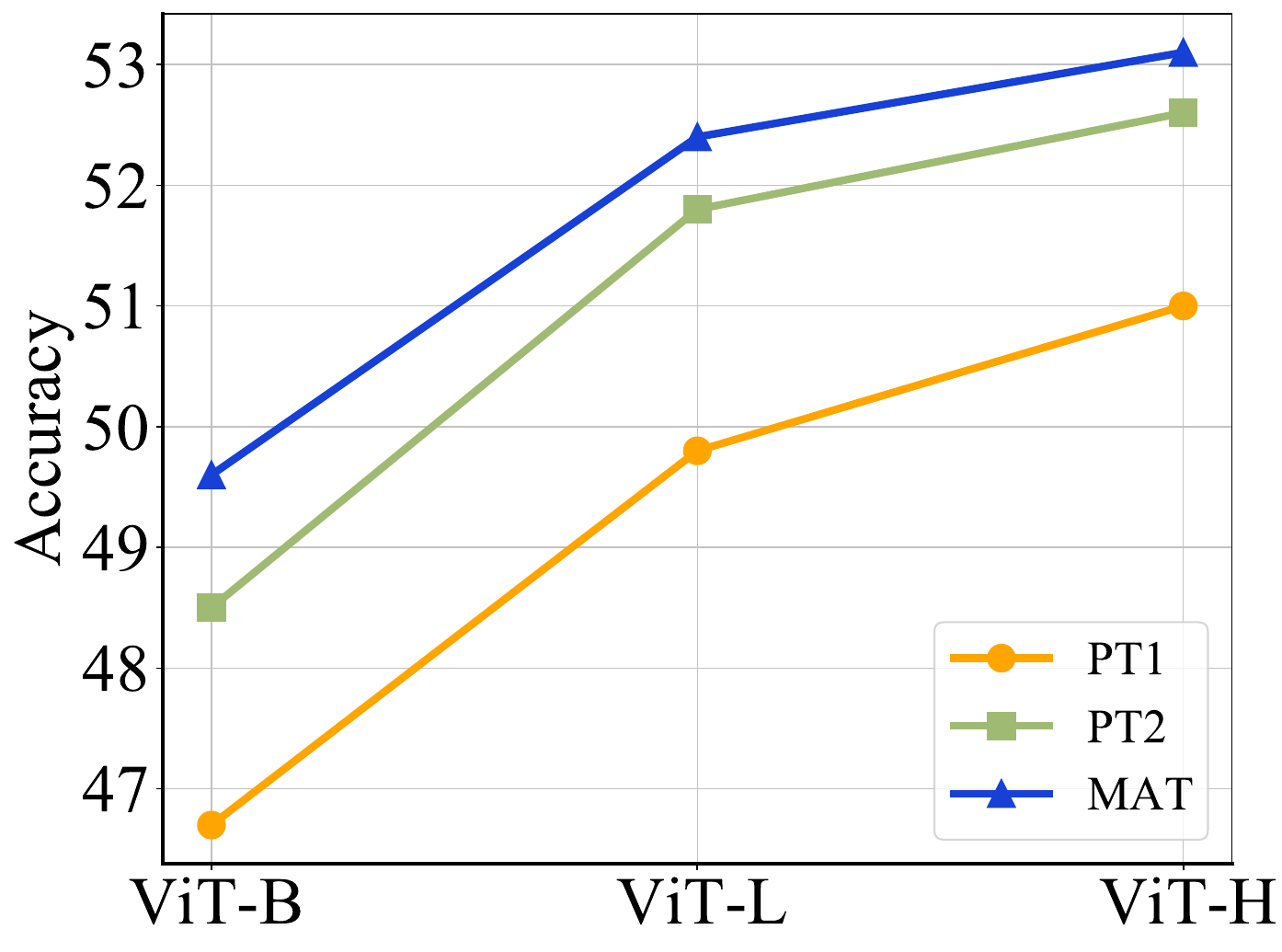}
  }
  \vspace{-.7em}
  \caption{Ablation experiments on various visual encoders. We report results on MSRVTT-QA and AVSD with ViT-B/16, ViT-L/14 and ViT-H/14.}
  \vspace{-.5em}
  \label{fig:visual_backbones}
\end{figure}

\begin{figure}[t]
  \centering
  \subcaptionbox{MSRVTT-QA}{
    \includegraphics[width=.47\linewidth]{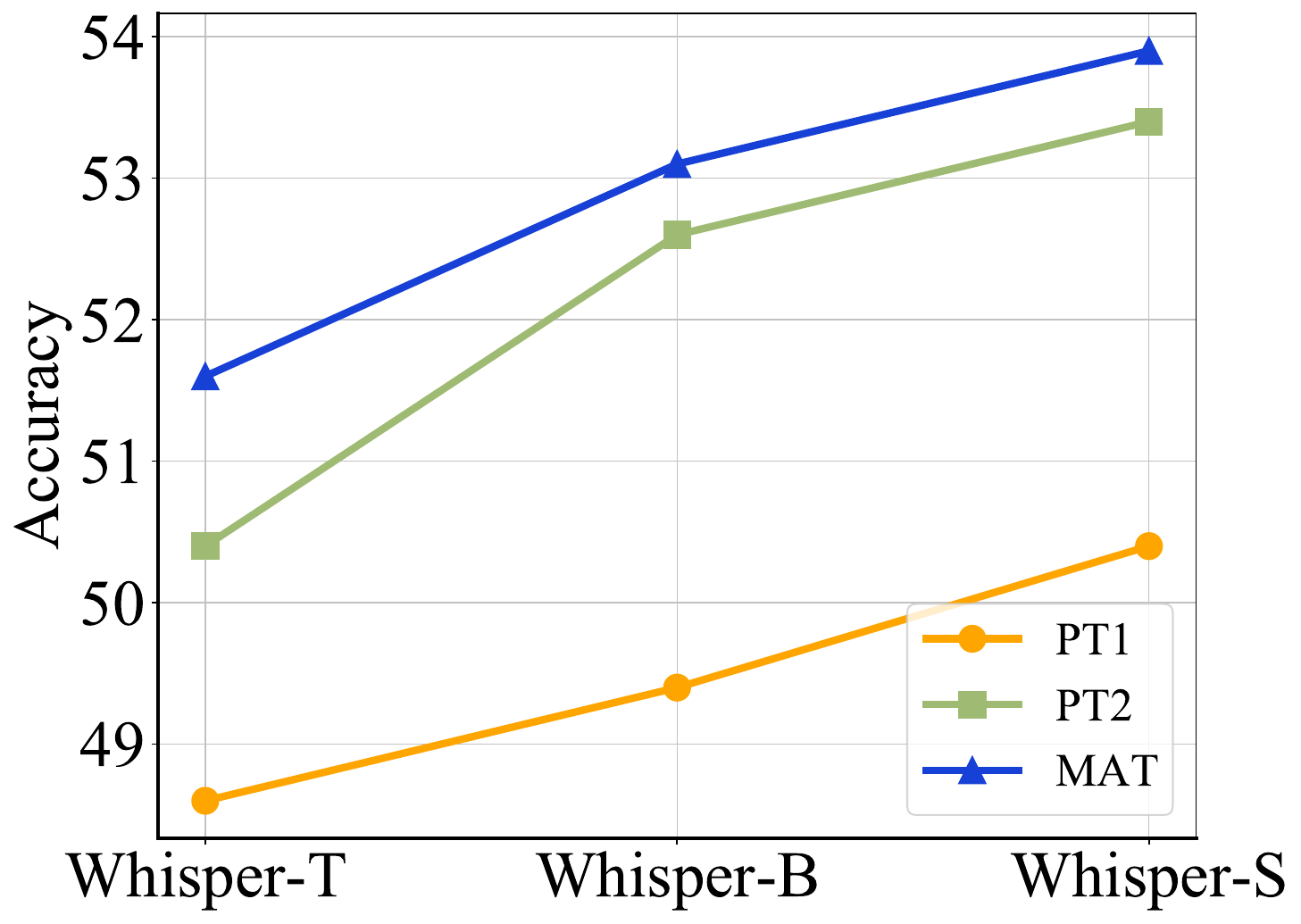}
  }
  \subcaptionbox{AVSD}{
    \includegraphics[width=.47\linewidth]{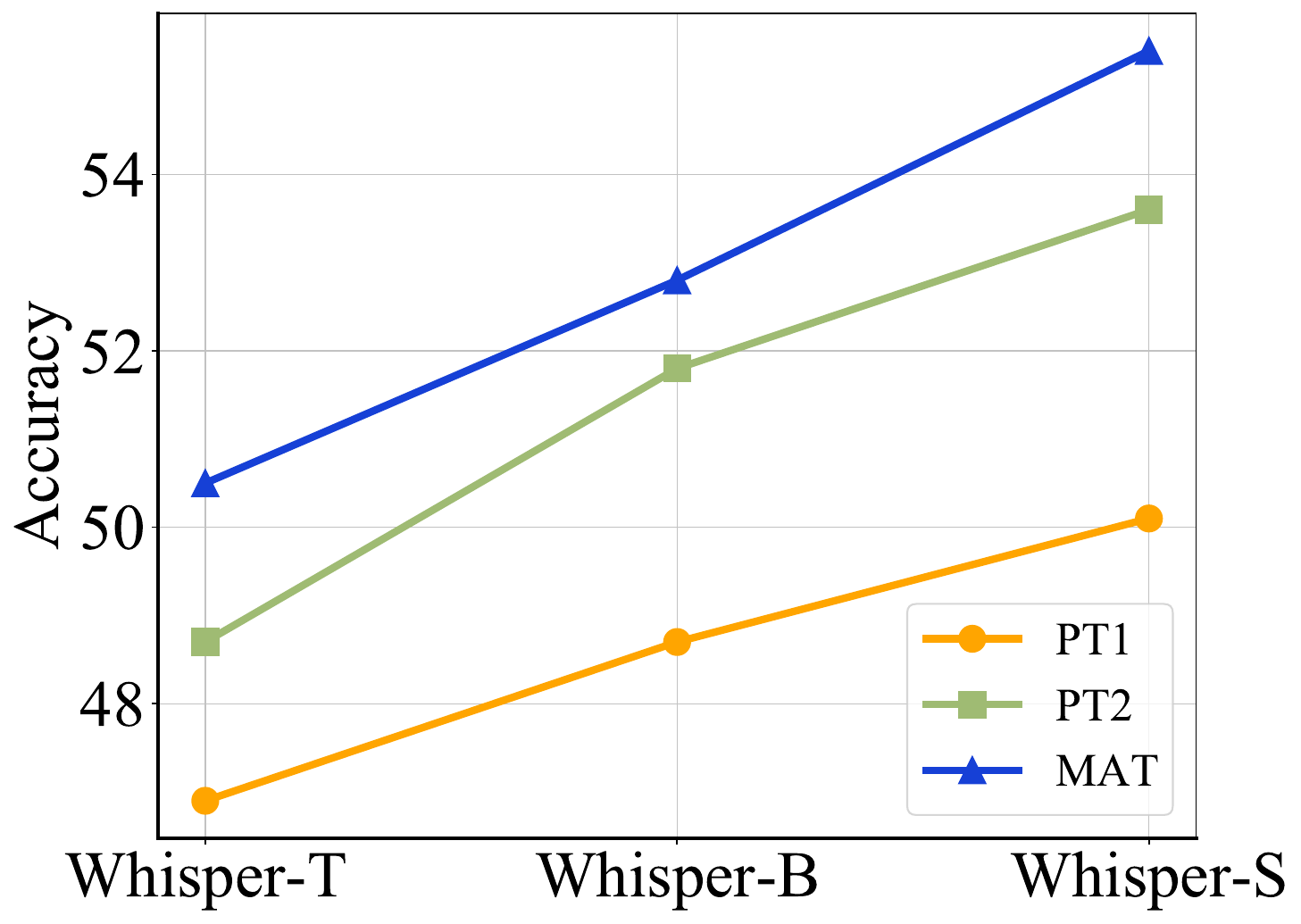}
  }
  \vspace{-.7em}
  \caption{Ablation experiments on various audio encoders. We report results on MSRVTT-QA and AVSD with Whisper-Tiny, Whisper-Base, and Whisper-Small.}
  \vspace{-.5em}
  \label{fig:audio_backbones}
\end{figure}

\begin{figure}[t]
  \centering
  \subcaptionbox{MSRVTT-QA}{
    \includegraphics[width=.47\linewidth]{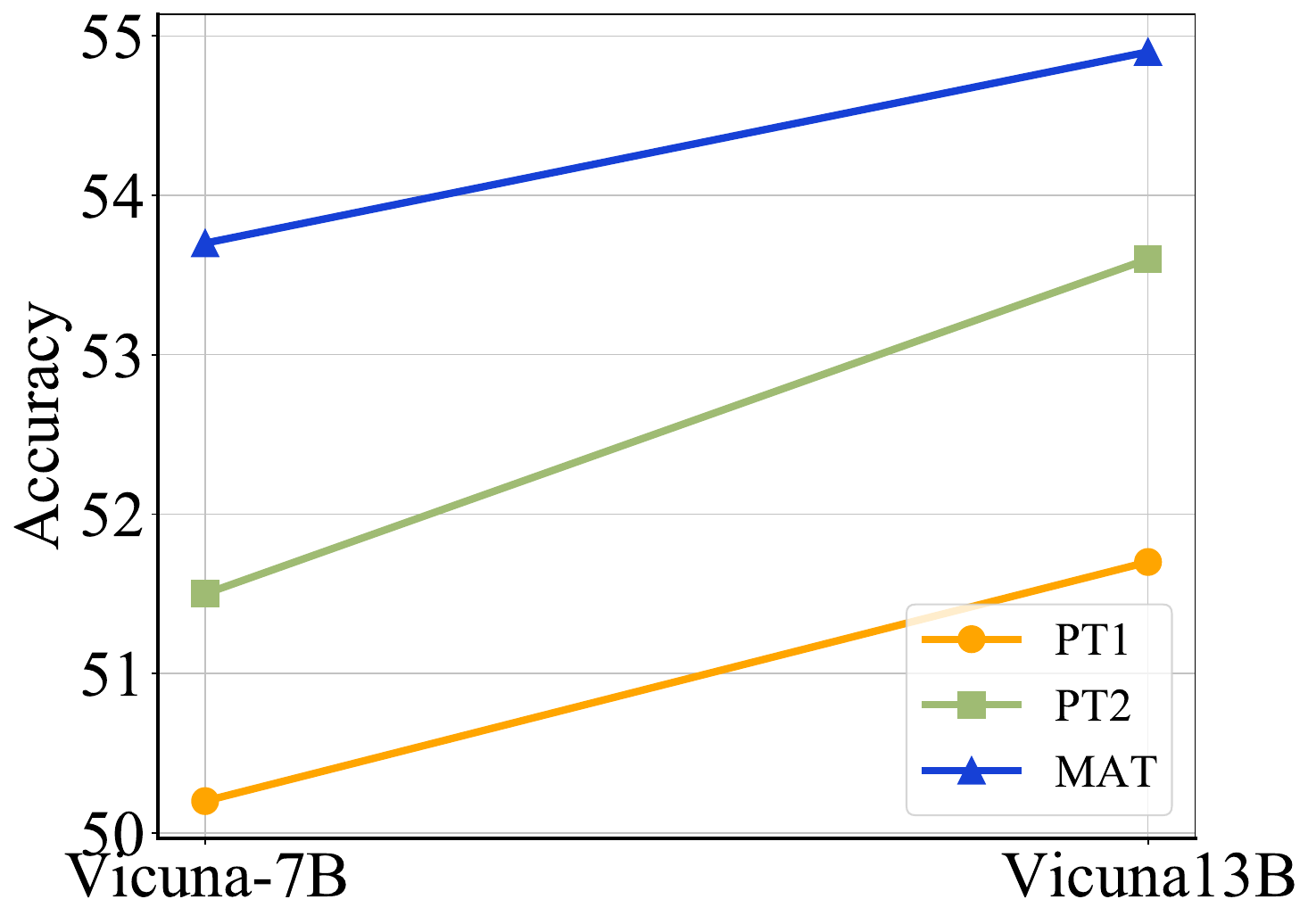}
  }
  \subcaptionbox{AVSD}{
    \includegraphics[width=.47\linewidth]{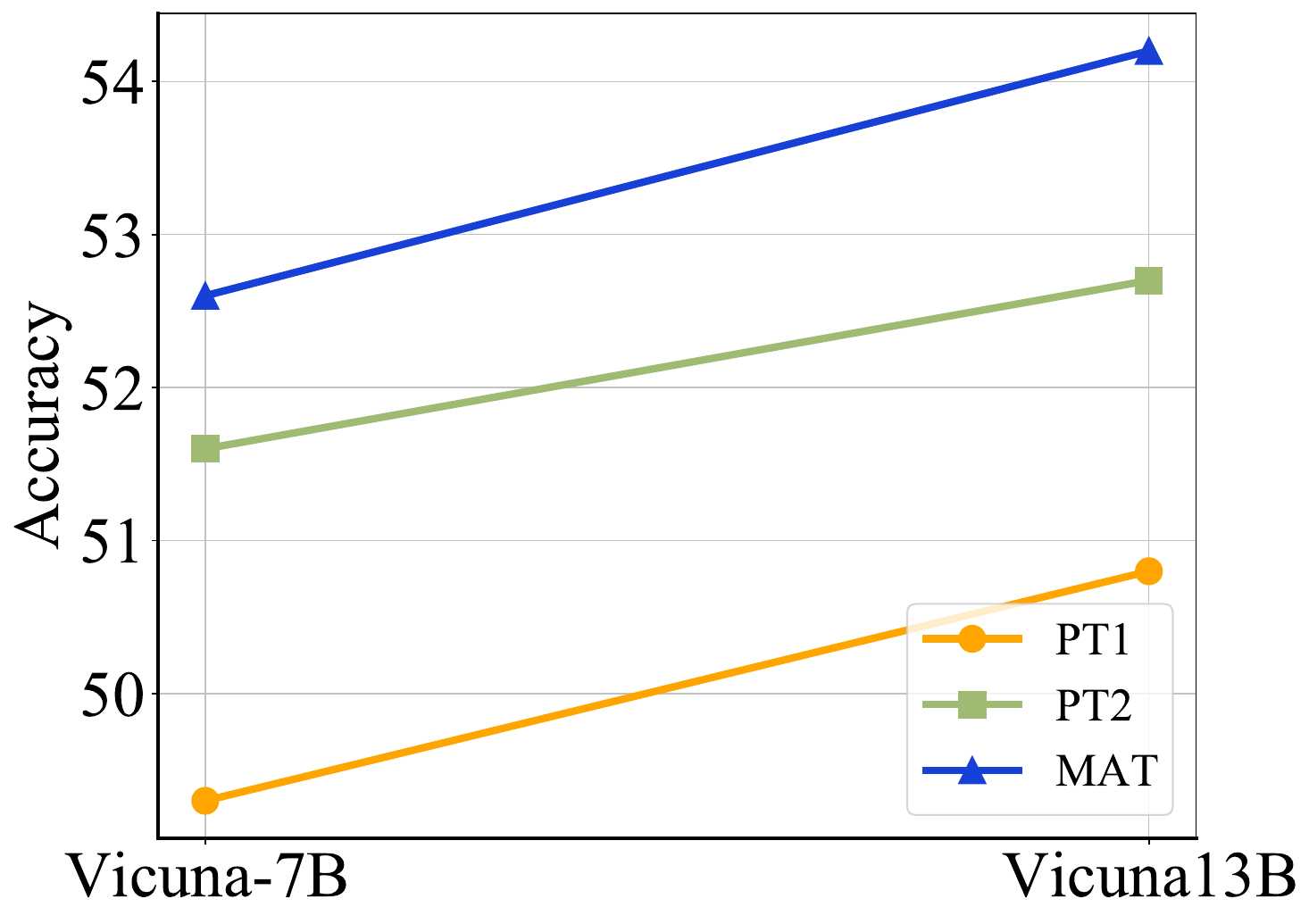}
  }
  \vspace{-.7em}
  \caption{Ablation experiments on LLMs. We report results on MSRVTT-QA and AVSD with Vicuna-7B and Vicuna-13B.}
  \label{fig:llm_backbones}
  \vspace{-1em}
\end{figure}

\begin{table}[t]
\centering
\scalebox{0.9}{
\begin{tabular}{l|ccc}
\toprule
\textbf{Method}  & \textbf{MSVD-QA}   & \textbf{MSRVTT-QA}   & \textbf{ActivityNet-QA} \\
\midrule
{PT1}  & {64.4}  & {50.2} & {44.7}  \\
{PT2}  & {65.9}  & {51.5}  & {45.6}  \\
\rowcolor{mygray-bg}{MAT}   & {67.3}  & {53.7}  & {47.2}  \\
\bottomrule
\end{tabular}}
\vspace{-.7em}
\caption{Ablation experiments on the training strategy. We report comparison results between modality-augmented training \textbf{(MAT)} versus the plain training of two versions: \textbf{PT1} and \textbf{PT2}.}
\label{tab:mat_pt}
\vspace{-2mm}
\end{table}

To evaluate the effectiveness of MAT, we conduct ablation experiments on the video QA tasks, including MSVD-QA, MSRVTT-QA, and ActivityNet-QA, using both visual and audio modalities from videos as input. We compare MAT with two versions of plain training: PT1 involves first joint training on the visual and audio branches with audio-visual data, and then separately training the visual/audio branch using visual/audio data, and PT2 involves first separate training on the visual/audio branch using visual/audio and then joint training on the visual and audio branches using audio-visual data.

\cref{tab:mat_pt} shows the results, where our MAT brings a +1.4\% on MSVD-QA, + 2.2\% MSRVTT-QA, and +1.6\% ActivityNet-QA than PT. Performance improvement indicates that our MAT can enhance multimodal video understanding compared to PT. Additionally, we observe that PT1 performs the worst. This is attributed to separate training on visual and audio branches, leading to bias in cross-modal alignment. Therefore, training with a mixture of modal samples achieves better cross-modal alignment in videos.

\begin{figure}[t]
    \centering
    \includegraphics[width=0.85\linewidth]{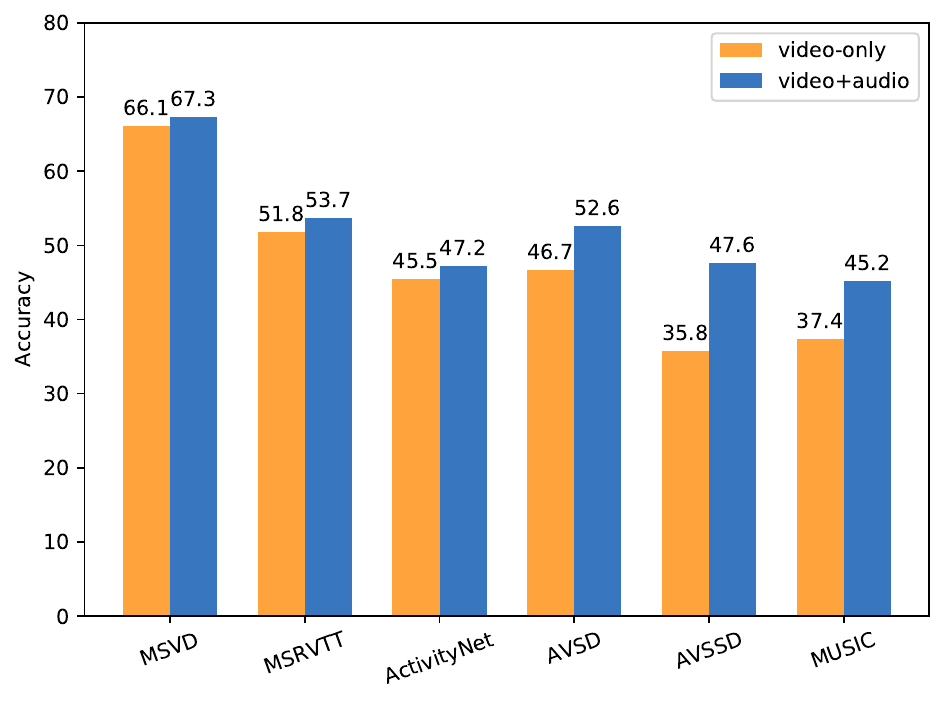}
    \vspace{-1.5em}
    \caption{Ablation experiments on integrating video modalities for video understanding. We report the comparison results on video QA and audio-visual QA between joint audio-visual modalities versus only the visual modality.}
    \vspace{-1.4em}
    \label{fig:modality_intrgration}
\end{figure}

We also report the comparative results on MSRVTT-QA~\cite{msrvtt} and AVSD~\cite{avsd} across visual encoders, audio encoders, and large language models (LLMs) of diverse scales. For visual encoders, we select ViT-B/16, ViT-L/14, and ViT-H/14. Considering that CLAP has only one size of HTSAT for audio encoders, we select Whisper~\cite{whisper} with Tiny, Base, and Small. For LLMs, we select Vicuna-7B and Vicuna-13B. As shown from \cref{fig:visual_backbones} to \cref{fig:llm_backbones}, Modality-Augmented Training (MAT) consistently outperforms the non-end-to-end single-modality Plain Training (PT) across various model architectures, indicating that MAT is not dependent on the specific network structure and possesses a strong generalizability.

\paragraph{Modality Integration.} Both visual and audio modalities contribute significantly to videos. To evaluate whether our method integrates visual and audio modalities to enhance the video understanding capability of LLMs. We conduct ablation experiments on video QA and audio-visual QA to show the necessity of joint audio-visual learning in video understanding. As illustrated in ~\cref{fig:modality_intrgration}, integrating both visual and auditory modalities, instead of relying on a single modality, consistently enhances performance across various video understanding benchmarks. These results highlight the importance of integrating visual elements and audio information within videos to provide a richer and more detailed understanding for video content. Our MAT with the well-curated audio-visual instruction dataset, captures the interactions and complementary information between different modalities, thereby enhancing the model's ability to interpret videos.


\begin{table}[t]
    \centering
    \scalebox{0.9}{ 
    \begin{tabular}{l|ccc}
    \toprule
    \textbf{{Vis Enc}} & \textbf{{MSVD-QA}} & \textbf{{MSRVTT-QA}}  & \textbf{{ActivityNet-QA}} \\
    \midrule
    {ViT-B/16} & {65.7}& {51.4} & {45.9} \\
    {ViT-L/14} & {67.3}& {53.7} & {47.2} \\
    \rowcolor{mygray-bg}{ViT-H/14} & {67.5}  & {54.2} & {47.5} \\
    \bottomrule
    \end{tabular}}
    \vspace{-.7em}
    \caption{Ablation experiments on scaling visual encoders. We report the comparisons among ViT-B/16, ViT-L/14, and ViT-H/14.}
    \label{tab:venc_scaling}
\end{table}

\begin{table}[t]
    \centering
    \scalebox{0.83}{ 
    \begin{tabular}{l|ccc}
    \toprule
    \textbf{{Aud Enc}} & \textbf{{MSVD-QA}} & \textbf{{MSRVTT-QA}}  & \textbf{{ActivityNet-QA}} \\
    \midrule
    {Whisper-Tiny} & {66.5}& {51.6} & {46.2} \\
    {Whisper-Base} & {67.7}& {53.1} & {47.4} \\
    \rowcolor{mygray-bg}{Whisper-Small} & {68.1}& {53.9} & {47.6} \\
    \bottomrule
    \end{tabular}}
    \vspace{-.7em}
    \caption{Ablation experiments on scaling audio encoders. We report comparisons among Whisper-Tiny, Base, and Small.}
    \label{tab:audio_scaling}
\end{table}

\begin{table}[t]
    \centering
    \scalebox{0.85}{ 
    \begin{tabular}{l|ccc}
    \toprule
    \textbf{{LLM}} & \textbf{{MSVD-QA}} & \textbf{{MSRVTT-QA}}  & \textbf{{ActivityNet-QA}} \\
    \midrule
    {Vicuna-7B} & {67.3}& {53.7} & {47.2} \\
    \rowcolor{mygray-bg}{Vicuna-13B} & {68.0}& {54.9} & {48.6} \\
    \bottomrule
    \end{tabular}}
    \vspace{-.7em}
    \caption{Ablation experiments on scaling LLMs. We report comparisons between Vicuna-7B and Vicuna-13B.}
    \label{tab:llm_scaling}
\end{table}

\begin{table}[t]
\centering
\scalebox{0.73}{
\begin{tabular}{l|ccccc}
\toprule
\textbf{Methods}  & \textbf{Correct}   & \textbf{Detail}  & \textbf{Context} & \textbf{Temporal} & \textbf{Consistency} \\
\midrule
{LLaMA-Adapter}  & {2.03}  & {2.32}  & {2.30} & {1.98} & {2.15} \\
{Video-LLaMA}  & {1.96} & {2.18} & {2.16}  & {1.82} & {1.79}  \\
{VideoChat}  & {2.23}  & {2.50}  & {2.53} & {1.94} & {2.24} \\
{Video-ChatGPT}  & {2.40}  & {2.52}  & {2.62} & {1.98} & {2.37} \\
{Valley}  & {2.43}  & {2.13}  & {2.86} & {2.04} & {2.45} \\
\rowcolor{mygray-bg}{Ours}  & {2.56}  & {2.47}  & {2.93} & {2.17} & {2.51} \\
\bottomrule
\end{tabular}}
\vspace{-.7em}
\caption{GPT-based evaluation on multiple dimensions for video understanding. We follow Video-ChatGPT to report score results (1$\sim$5) on correct, detail, context, temporal, and consistency.}
\label{tab:diverse_dimensions}
\vspace{-.5em}
\end{table}

\paragraph{Size of Model Architecture.} 
The size of the model architecture is a pivotal factor in influencing performance. In our method, we employ ViT-L/14 from CLIP, HTSAT from CLAP, and Vicuna-7B from LLaMA as the visual encoder, audio encoder, and LLMs, respectively. We conduct ablation experiments across various scales of the visual encoder, audio encoder, and LLMs components. 

As shown from \cref{tab:venc_scaling} to \cref{tab:llm_scaling}, we observe that increasing the size of the multimodal encoders and LLM backbone leads to performance improvements. The ablation experiments on model size also suggest that combining the powerful multimodal encoder with fine-grained representation capability and LLM with strong reasoning ability can achieve excellent performance.



\paragraph{Compare on Multiple Dimensions.}

The default evaluation focuses on a single dimension (accuracy). To fully evaluate the effectiveness of our method, we follow Video-ChatGPT and use GPT-4 to score (1$\sim$5) across multiple dimensions: Correctness of Information (Correct), Detail Orientation (Detail), Contextual Understanding (Context), Temporal Understanding (Temporal), and Consistency. The results in ~\cref{tab:diverse_dimensions} demonstrate that our method shows significant improvement over previous work, proving the efficacy of our method.

\paragraph{Length of Sequence.} 
In videos, each frame captures instantaneous visual information, while audio segments contain instantaneous sound information. Therefore, more video frames and audio segments can provide more details and contextual information, thereby assisting LLM in a more comprehensive understanding of video content. By default, we use 32 video frames and 4 audio segments. We conduct ablation experiments to explore the effect of the sequence length of video frames and audio segments. In the ablation of video frames, we experiment with various lengths in \{4,8,16,32,64\}. In the ablation of audio segments, we experiment with various lengths in \{1,2,4,8,16\}. 

\begin{figure}[t]
  \centering
  \subcaptionbox{ActivityNet-QA}{
    \includegraphics[width=.47\linewidth]{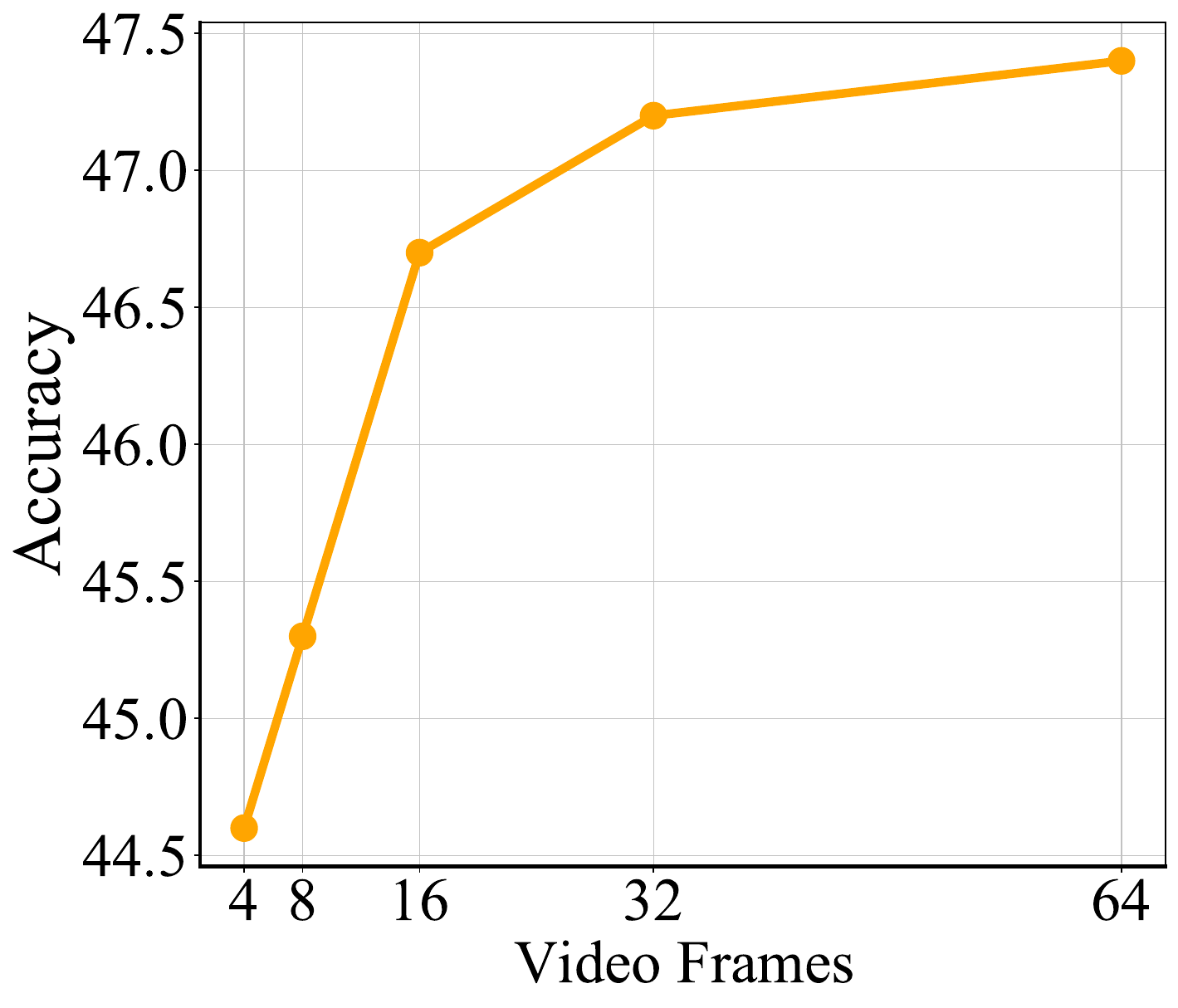}
  }\hfill
  \subcaptionbox{MSRVTT-QA}{
    \includegraphics[width=.47\linewidth]{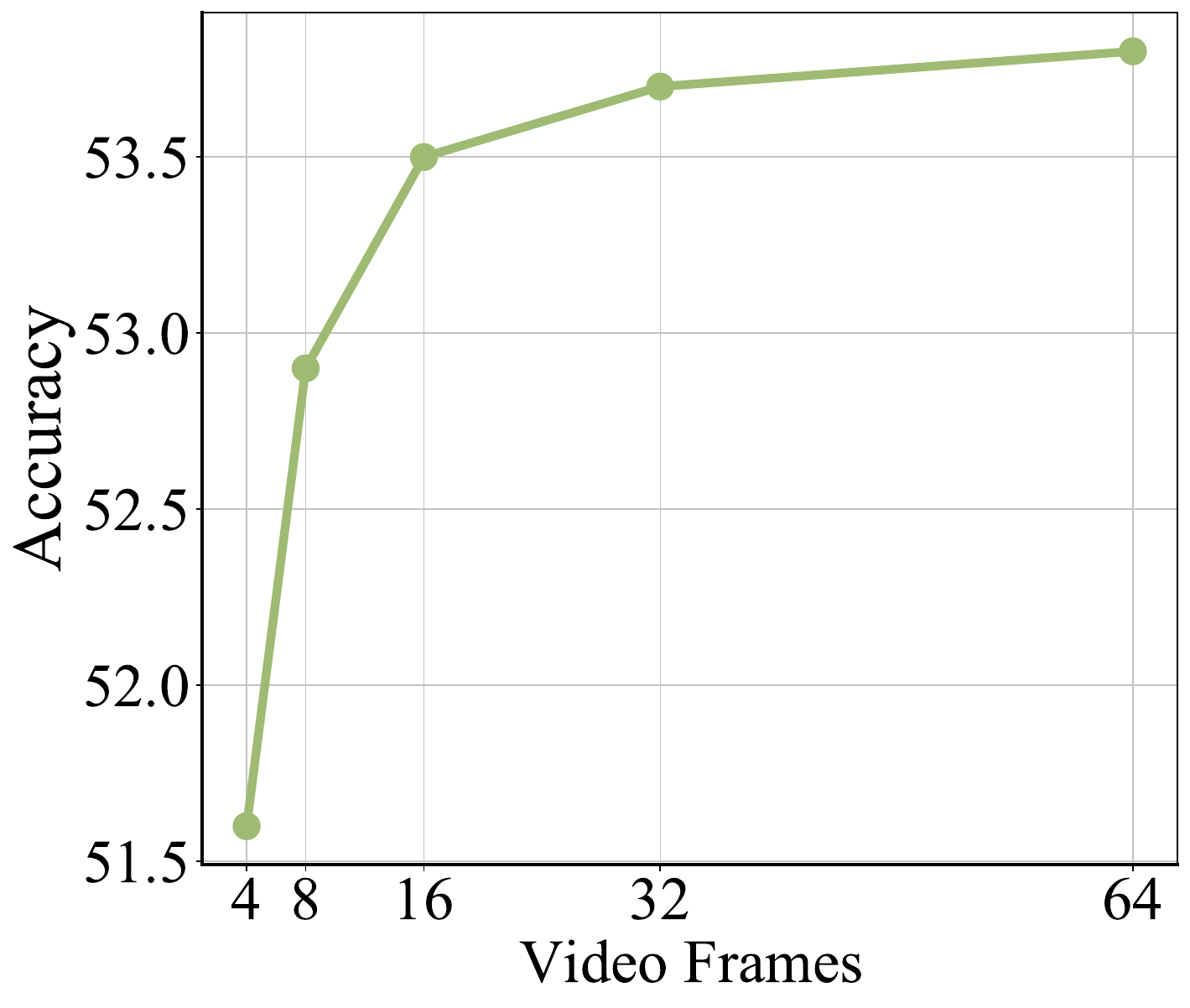}
  }
  \vspace{-.7em}
  \caption{Ablation experiments on lengths of video frames.  We report results on video frames of length 4, 8, 16, 32, and 64.}
  \label{fig:frame_scaling}
\end{figure}

\begin{figure}[t]
  \centering
  \subcaptionbox{ActivityNet-QA}{
    \includegraphics[width=.47\linewidth]{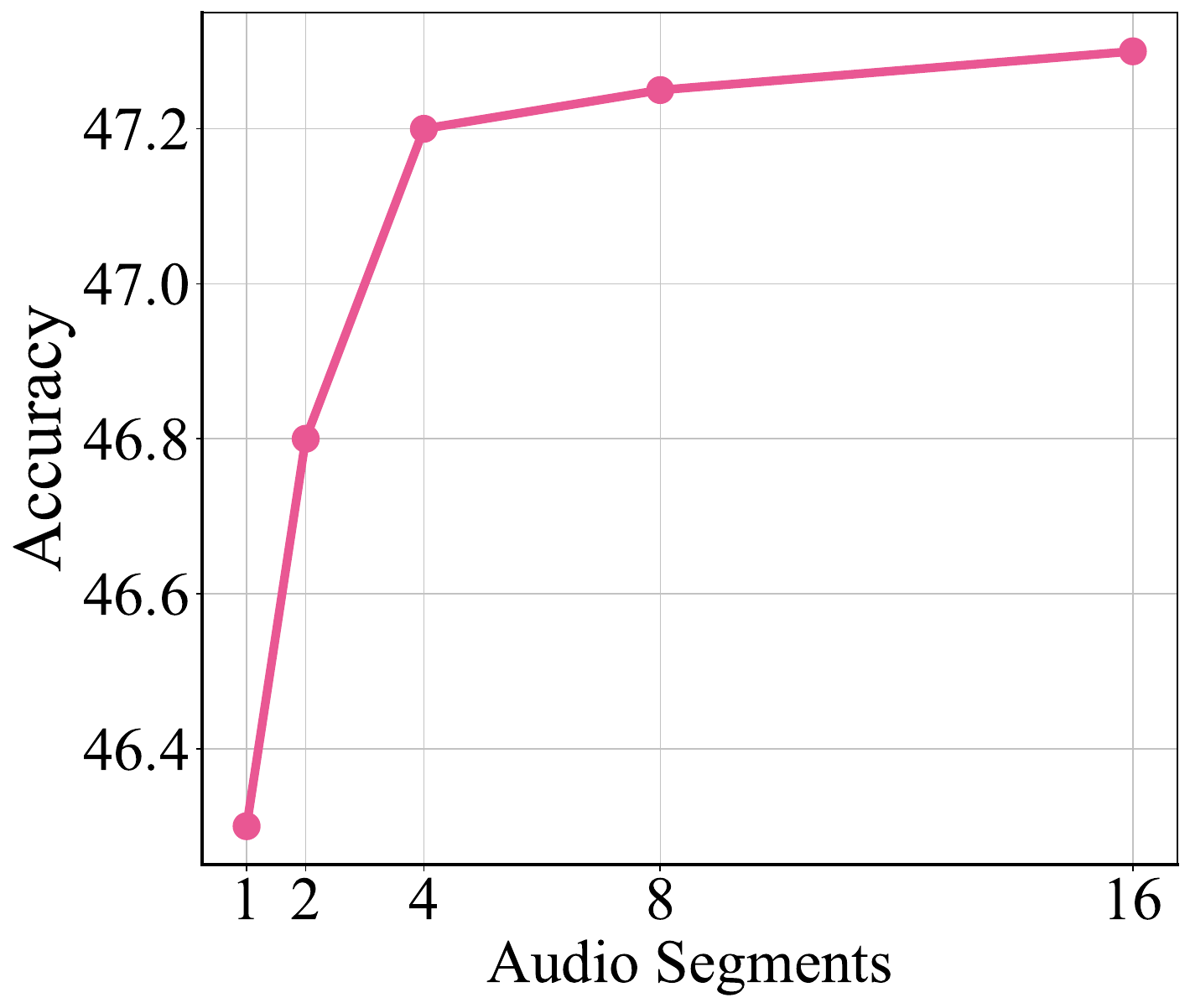}
  }\hfill
  \subcaptionbox{MSRVTT-QA}{
    \includegraphics[width=.47\linewidth]{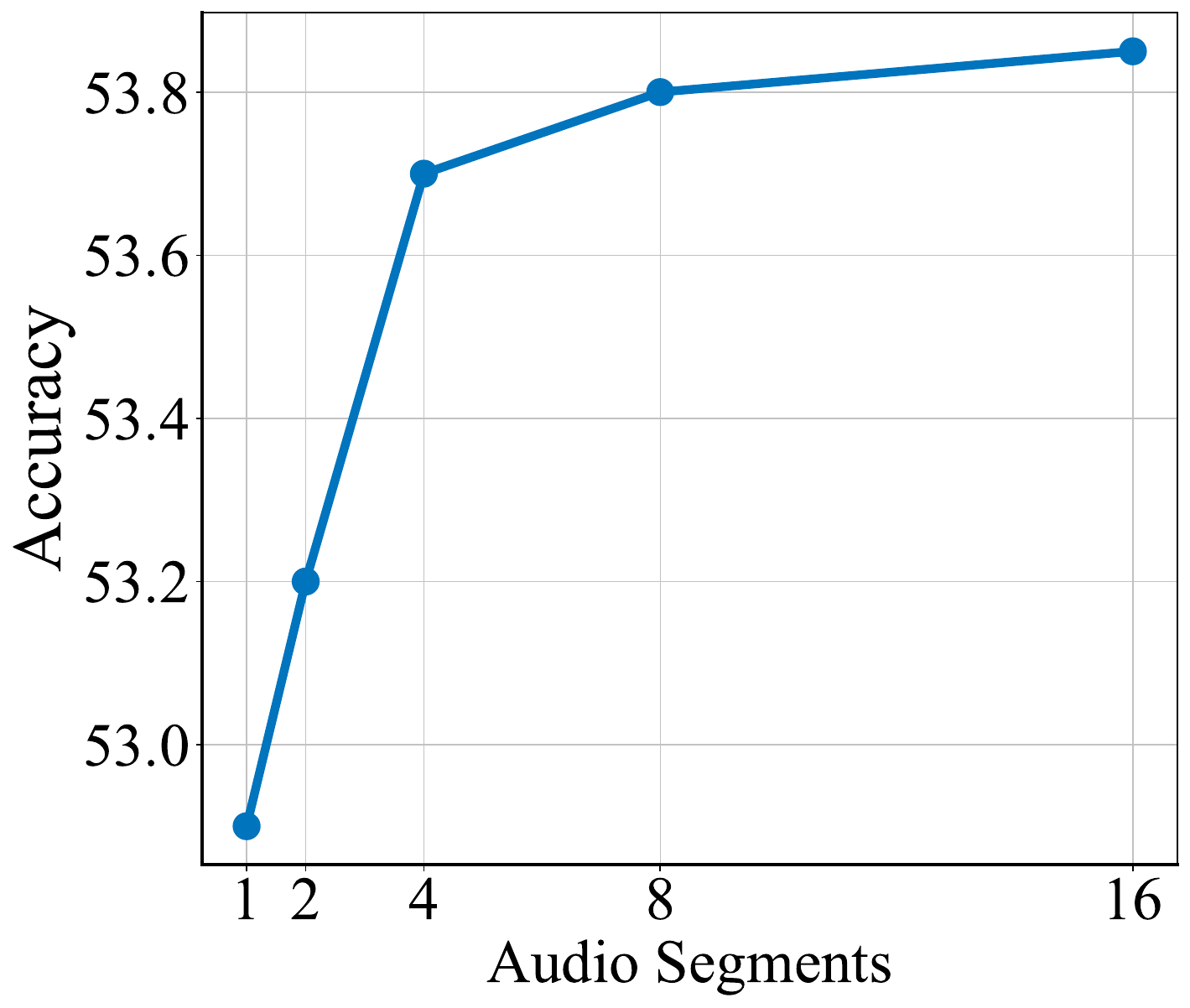}
  }
  \vspace{-.7em}
  \caption{Ablation experiments on lengths of audio segments. We report results on audio segments of length 1, 2, 4, 8, and 16.}
  \label{fig:segment_scaling}
  \vspace{-1em}
\end{figure}

\cref{fig:frame_scaling} and \cref{fig:segment_scaling} show the results on the video QA tasks with different average video durations, including MSRVTT-QA (15s) and ActivityNet-QA (180s). The accuracy curve demonstrates that increasing the sequence length does improve the model's performance. However, as the sequence length increases, the rate of performance improvement gradually diminishes. This suggests that longer sequences provide more information, but the redundancy in the video limits the extent to which performance can be further enhanced. Future research can explore the module to strike a balance between obtaining sufficient information from longer video sequences and reducing the computational burden caused by redundant data. This will enable LLM to efficiently focus on key video frames and audio segments.

\paragraph{Tuning Parameters}
The default configuration in our experiment is full tuning. This method requires a lot of training resources. Recent works explore freezing the model parameters and employing Low-Rank Adaptation (LoRA~\cite{lora}) to learn additional parameters for model fusion. We also compared LoRA with full tuning. The results in ~\cref{tab:tuning_params} show that LoRA can achieve comparable results with fewer GPU resources. However, there is still a gap compared to full tuning. We analyze that the limited number of parameters in LoRA may not be sufficient to store visual and audio information. Further research could investigate how to jointly train the visual encoder, audio encoder, and LLM with LoRA.

\begin{table}[h]
\centering
\scalebox{0.8}{
\begin{tabular}{l|ccc}
\toprule
\textbf{Tuning Params}  & \textbf{MSVD-QA}   & \textbf{MSRVTT-QA}  & \textbf{ActivityNet-QA} \\
\midrule
{LoRA}  & {62.8}  & {47.5} & {44.1}   \\
{Full Tuning}  & {67.3}  & {53.7}  & {47.2}  \\
\bottomrule
\end{tabular}}
  \vspace{-.7em}
\caption{Ablation experiments on the tuning parameters. We report results on video QA tasks between LoRA and full tuning.}
\label{tab:tuning_params}
\vspace{-.5em}
\end{table}
\section{Conclusion}
\label{sec:conclusion}

This paper introduces Audio-Visual LLM, a multimodal framework that empowers LLM with video instruction-following capability. The modality-augmented training plays a crucial role in enabling end-to-end joint training with video data across different modalities, including visual-only, audio-only, and audio-visual formats. Additionally, we present a high-quality video instruction dataset derived from GPT-4, which enables our model to effectively process a wide range of task-oriented video instructions, spanning from multi-turn conversations and audio-visual narratives to complex reasoning tasks. Extensive experiments demonstrate the impressive performance of Audio-Visual LLM across diverse video understanding tasks.

\section*{Supplementary Material}
\appendix


\section{Datasets}
\begin{figure*}[h]
    \centering
    \scalebox{0.98}{
    \includegraphics[width=1.0\linewidth]{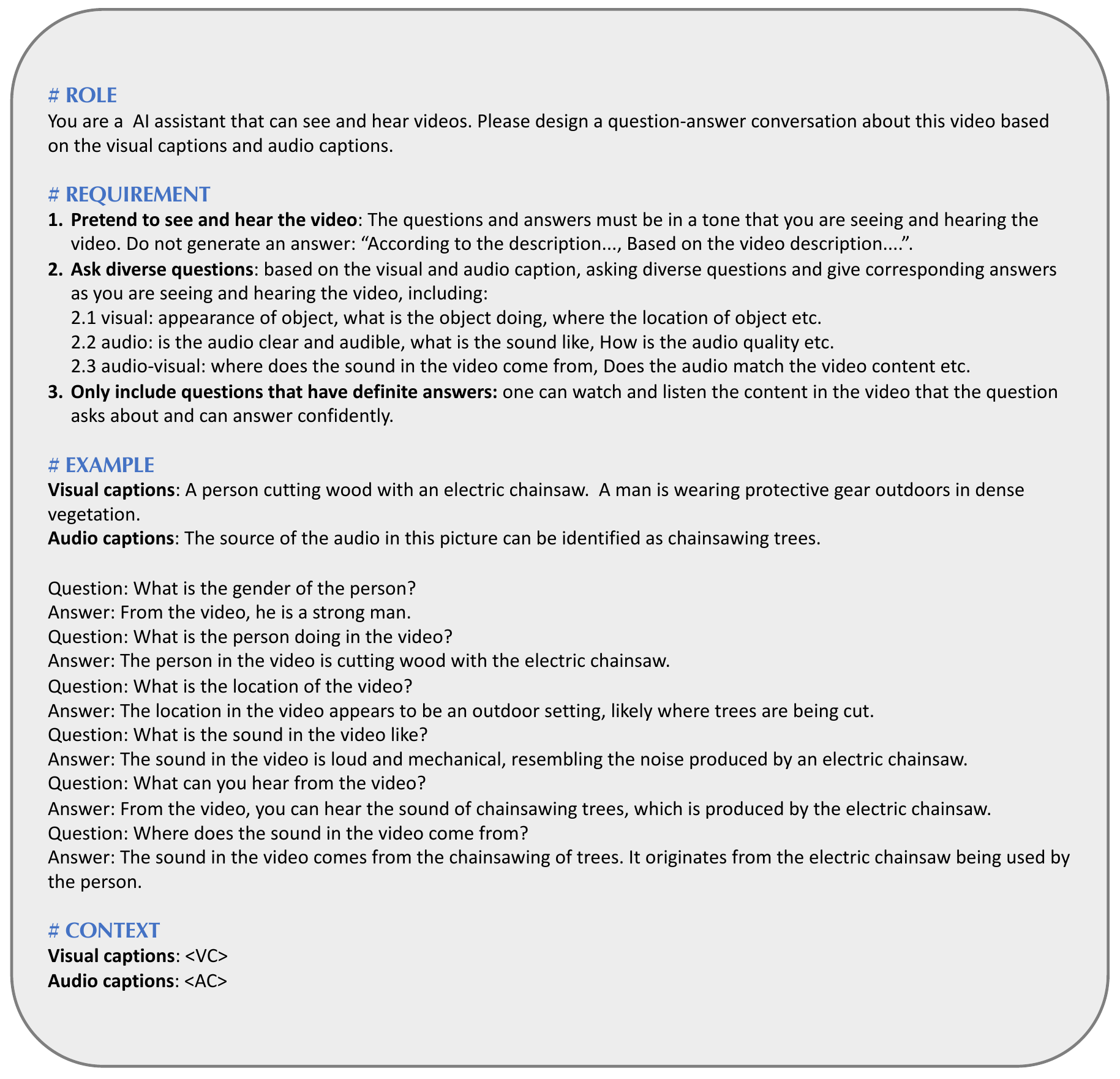}
    }
    \caption{Prompt template for multi-turn conversations.}
    \vspace{-1em}
    \label{fig:multi-turn conversation}
\end{figure*}

\begin{figure*}[h]
    \centering
    \scalebox{0.98}{
    \includegraphics[width=1.0\linewidth]{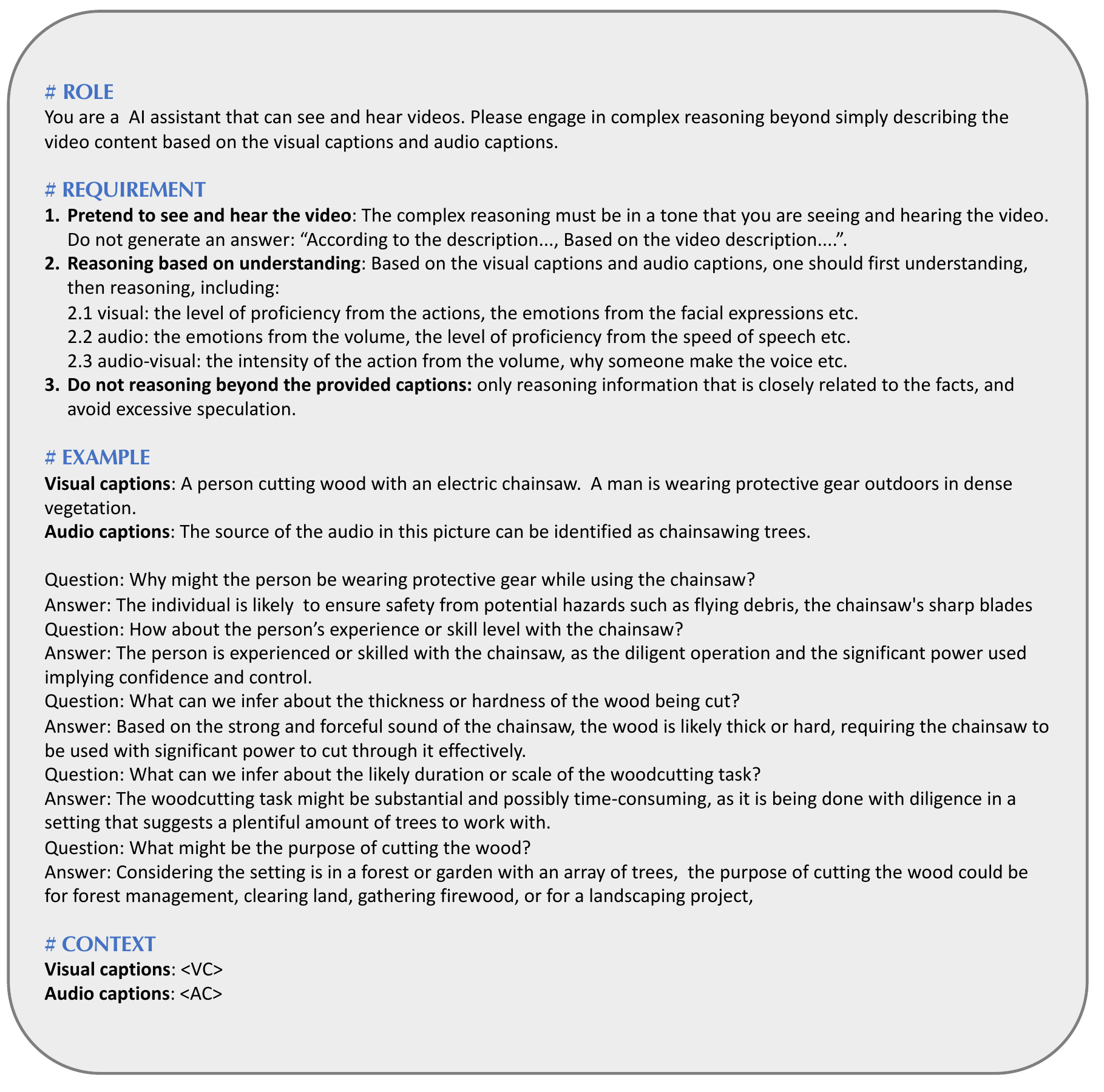}
    }
    \caption{Prompt template for complex reasoning.}
    \vspace{-1em}
    \label{fig:complex reasoning}
\end{figure*}

\begin{figure*}[h]
    \centering
    \scalebox{0.98}{
    \includegraphics[width=1.0\linewidth]{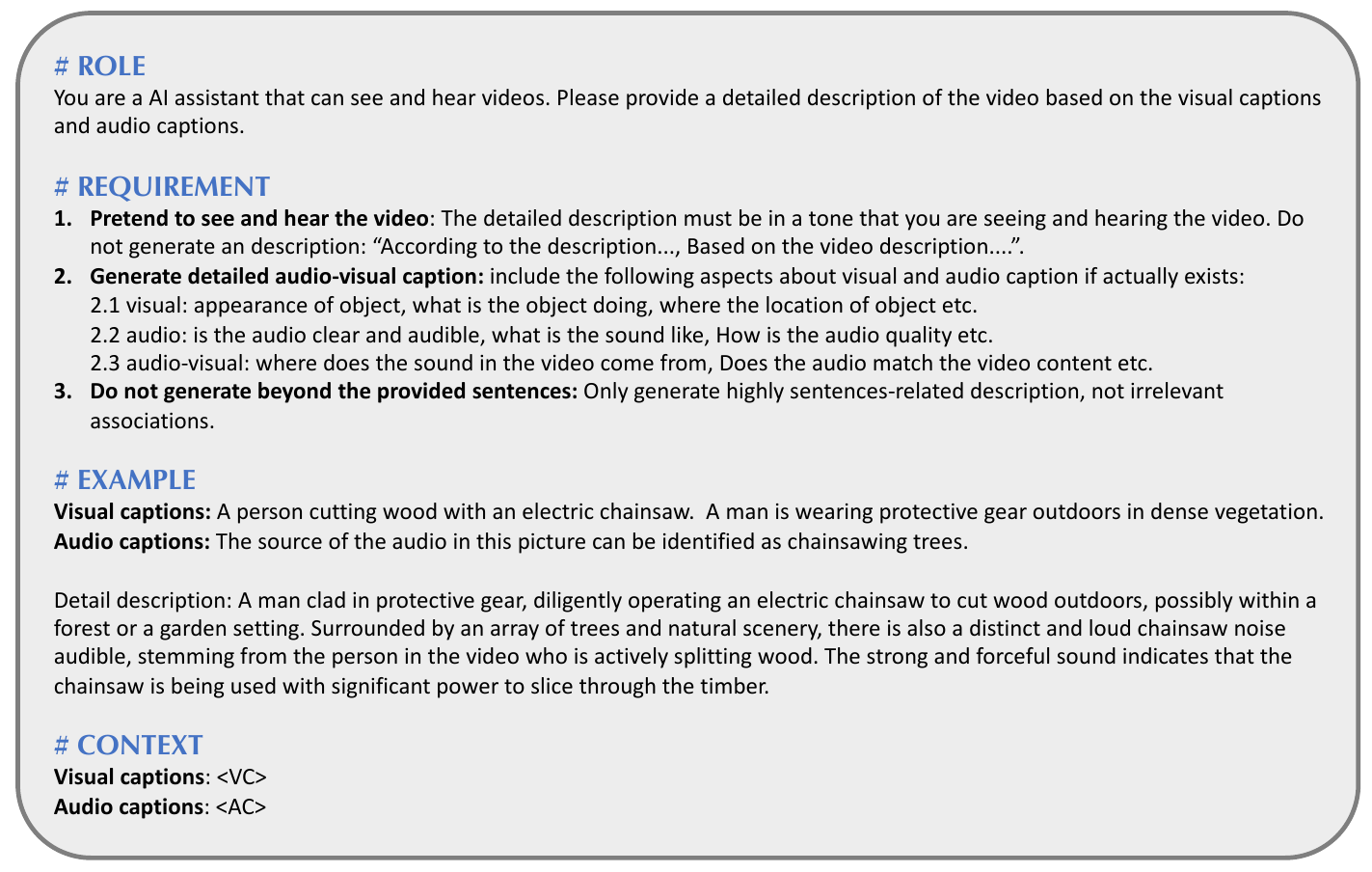}
    }
    \caption{Prompt template for audio-visual description.}
    \vspace{-1em}
    \label{fig:audio-visual description}
\end{figure*}

We provide a detailed introduction to the structure of the prompt templates, where ~\cref{fig:multi-turn conversation} presents the multi-turn conversations, ~\cref{fig:complex reasoning} presents the complex reasoning, and ~\cref{fig:audio-visual description} presents the detailed audio-visual descriptions.

The prompt template is structured to include \textcolor{blue}{ROLE}, \textcolor{blue}{REQUIREMENT}, \textcolor{blue}{EXAMPLE}, and \textcolor{blue}{CONTEXT}. The \textcolor{blue}{ROLE} defines the task for GPT-4, and the \textcolor{blue}{REQUIREMENT} outlines the specific criteria for instruction generation. The \textcolor{blue}{EXAMPLE} provides a prototypical sample for GPT-4's reference. In the \textcolor{blue}{CONTEXT}, $\texttt{<VC>}$ and $\texttt{<AC>}$ are substituted with the provided visual and audio captions. This approach enables converting video content into textural descriptions that GPT-4 can understand, facilitating the generation of instructive video data. As shown in ~\cref{tab:data_detail}, we produce 260k instruction data pairs from audio, visual, and audio-visual samples, encompassing 100k detailed audio-visual descriptions, 120k multi-turn conversations, and 40k complex reasoning.

\begin{table}[h]
\centering
\scalebox{0.88}{
\begin{tabular}{l|cccc}
\toprule
\textbf{Instruction Type}  & \textbf{Audio}   & \textbf{Visual}  & \textbf{Aud-Vis} & \textbf{Total}\\
\midrule
{Multi-Turn Conversation}  & {20}  & {60} & {40}  & {120} \\
{Detailed Description}   & {20}  & {50}  & {30} & {100} \\
{Complex Reasoning}  & {5}  & {20}  & {15} & {40} \\
\bottomrule
\end{tabular}}
\caption{The data distribution of the instruction dataset. We generate various instruction types including multi-turn conversations,  detailed descriptions, and complex reasoning. We also cover various video data types, including audio, visual, and audio-visual samples.}
\label{tab:data_detail}
\end{table}

\section{Additional Experiments}


\subsection{Effect of Modality-Augmented Training}
The default model architecture configurations are ViT-L/14 (CLIP~\cite{clip}), HTSAT (CLAP~\cite{clap}), Vicuna-7B (LLaMA~\cite{llama}). We conduct additional experiments on video QA and audio-visual QA to verify the effectiveness of our Modality-Augmented Training (MAT) across visual encoders, audio encoders, and large language models (LLMs) of diverse scales. For visual encoders, we select ViT-B/16, ViT-L/14, and ViT-H/14. Considering that CLAP has only one size of HTSAT for audio encoders, we select Whisper~\cite{whisper} with Tiny, Base, and Small. For LLMs, we select Vicuna-7B and Vicuna-13B.

We report the comparative results on MSRVTT-QA~\cite{msrvtt} and AVSD~\cite{avsd}. As shown from \cref{fig:visual_backbones} to \cref{fig:llm_backbones}, Modality-Augmented Training (MAT) consistently outperforms the non-end-to-end single-modality Plain Training (PT) across various model architectures, indicating that the joint learning of visual and audio modalities indeed helps the model to understand videos comprehensively. Moreover. our MAT is not dependent on the specific network structure and possesses a strong generalizability.

\begin{figure}[htbp]
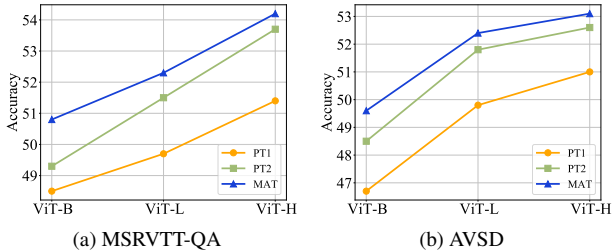

  \centering
  \subcaptionbox{MSRVTT-QA}{
    \includegraphics[width=.47\linewidth]{figs/visual_backbones_msrvtt.pdf}
  }
  \subcaptionbox{AVSD}{
    \includegraphics[width=.47\linewidth]{figs/visual_backbones_avsd.pdf}
  }
  \caption{Ablation experiments on various visual encoders. We report results on MSRVTT-QA and AVSD with ViT-B/16, ViT-L/14 and ViT-H/14.}
  \label{fig:visual_backbones}
\end{figure}

\begin{figure}[htbp]
  \centering
  \subcaptionbox{MSRVTT-QA}{
    \includegraphics[width=.47\linewidth]{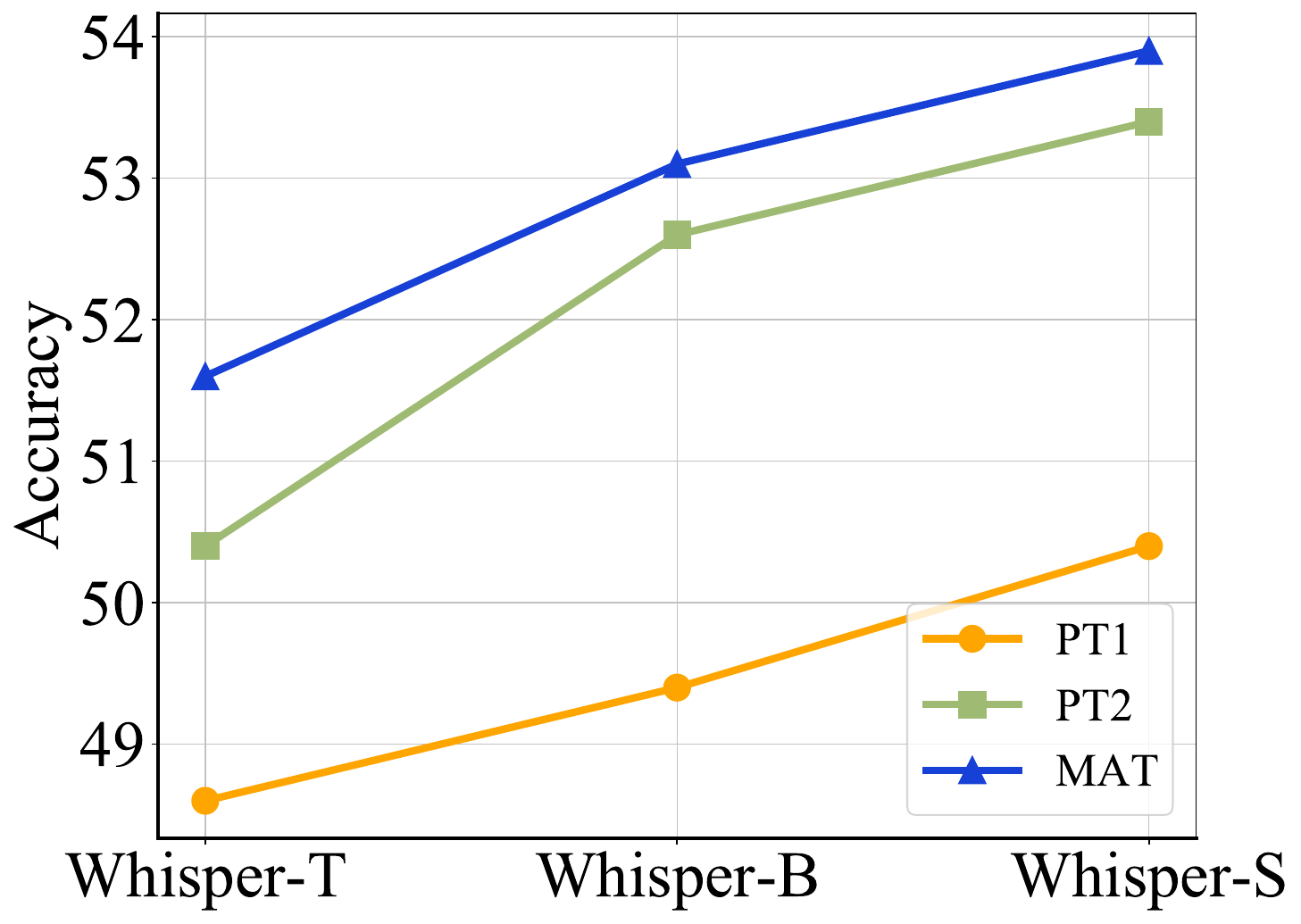}
  }
  \subcaptionbox{AVSD}{
    \includegraphics[width=.47\linewidth]{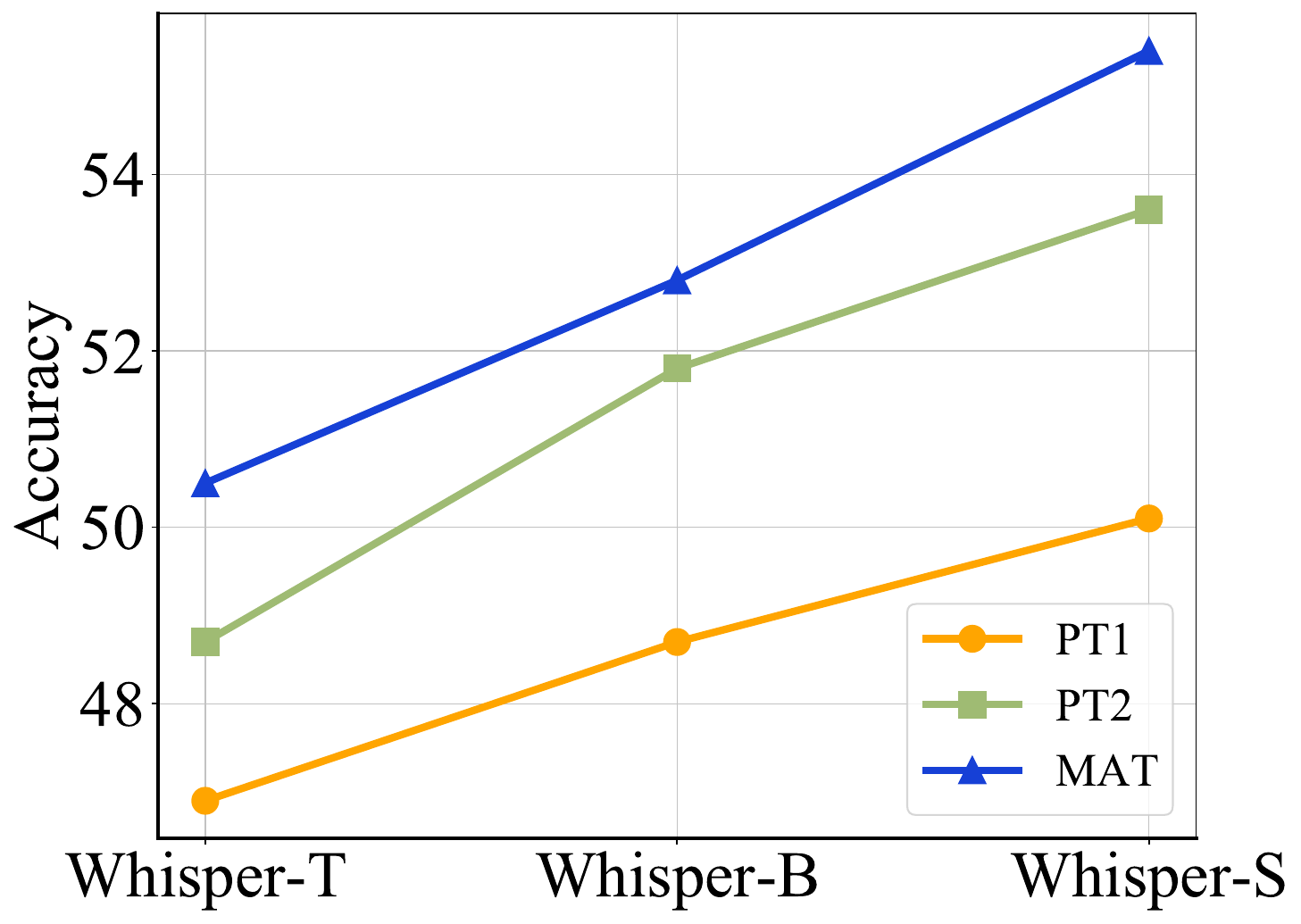}
  }
  \caption{Ablation experiments on various audio encoders. We report results on MSRVTT-QA and AVSD with Whisper-Tiny, Whisper-Base, and Whisper-Small.}
  \label{fig:audio_backbones}
\end{figure}

\begin{figure}[htbp]
  \centering
  \subcaptionbox{MSRVTT-QA}{
    \includegraphics[width=.47\linewidth]{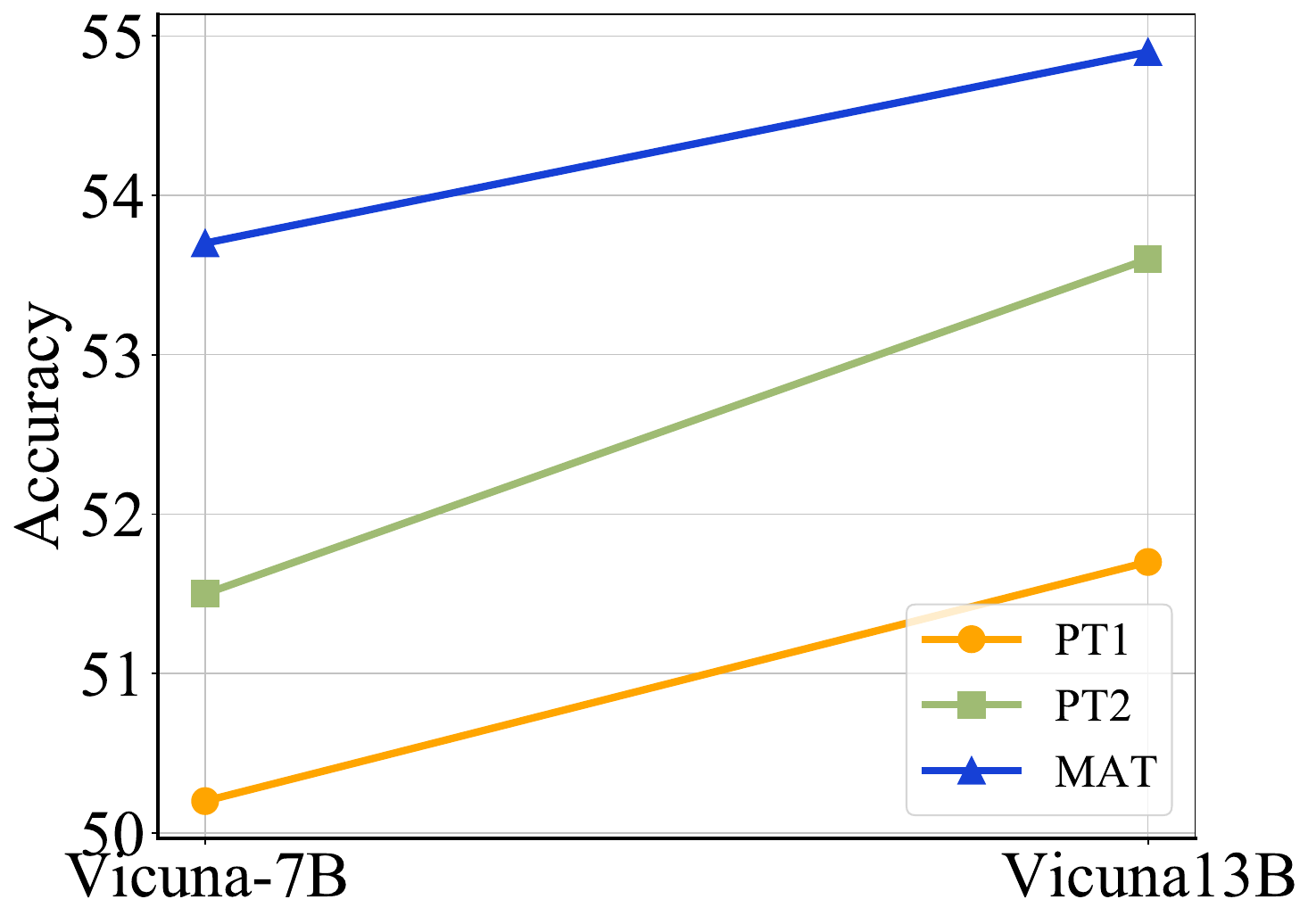}
  }
  \subcaptionbox{AVSD}{
    \includegraphics[width=.47\linewidth]{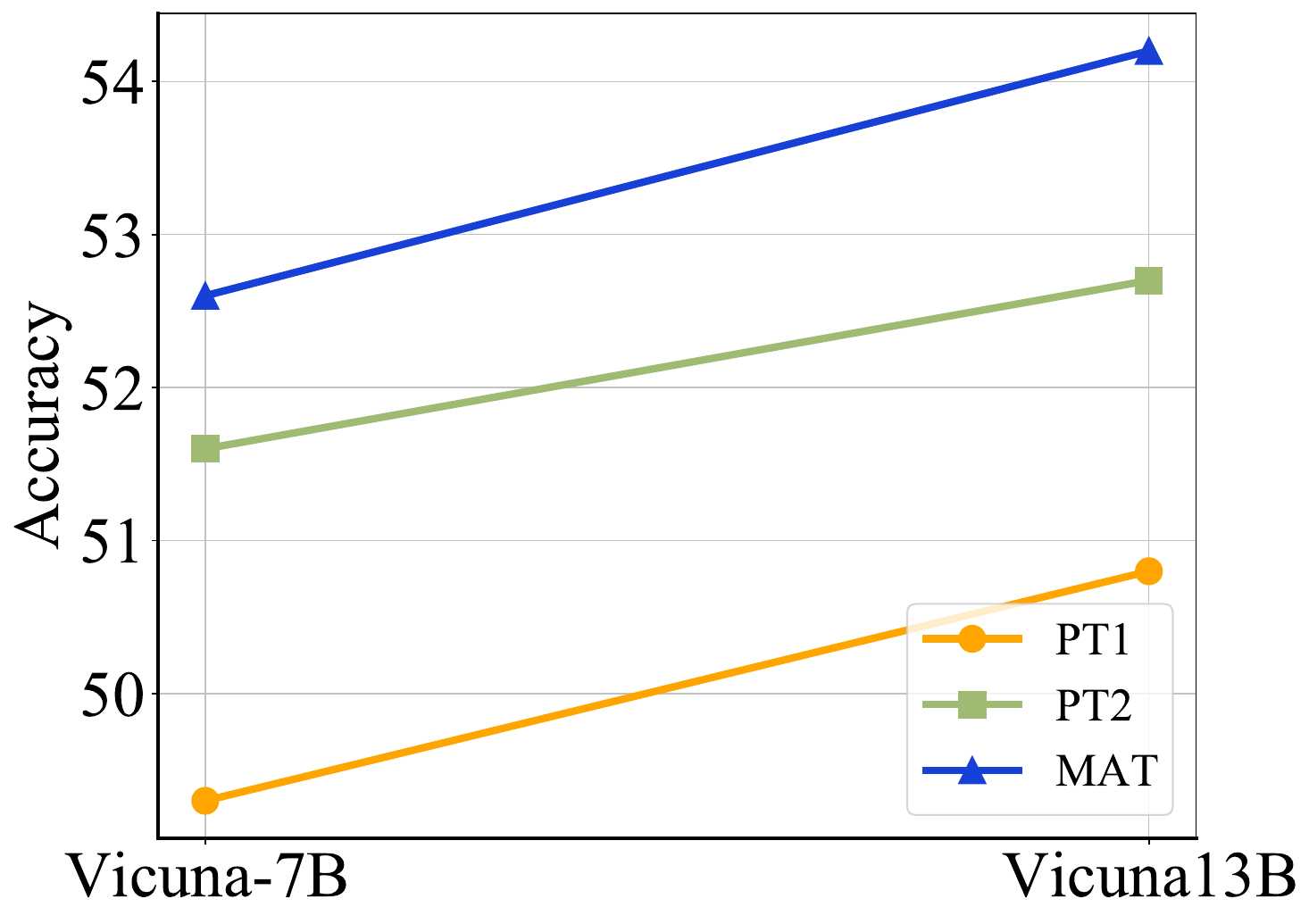}
  }
  \caption{Ablation experiments on various LLMs. We report results on MSRVTT-QA and AVSD with Vicuna-7B and Vicuna-13B.}
  \label{fig:llm_backbones}
\end{figure}

\subsection{Effect of Modality Integration}

We conduct additional experiments on video QA and audio-visual QA to show the necessity of joint audio-visual learning in video understanding. As illustrated in ~\cref{fig:modality_intrgration}, integrating both visual and auditory modalities, instead of relying on a single modality, consistently enhances performance across various video understanding benchmarks. These results highlight the importance of integrating visual elements and audio information within videos to provide a richer and more detailed understanding for video content. Our MAT with the well-curated audio-visual instruction dataset, captures the interactions and complementary information between different modalities, thereby enhancing the model's ability to interpret videos.

\begin{figure}[!ht]
    \centering
    \includegraphics[width=1.0\linewidth]{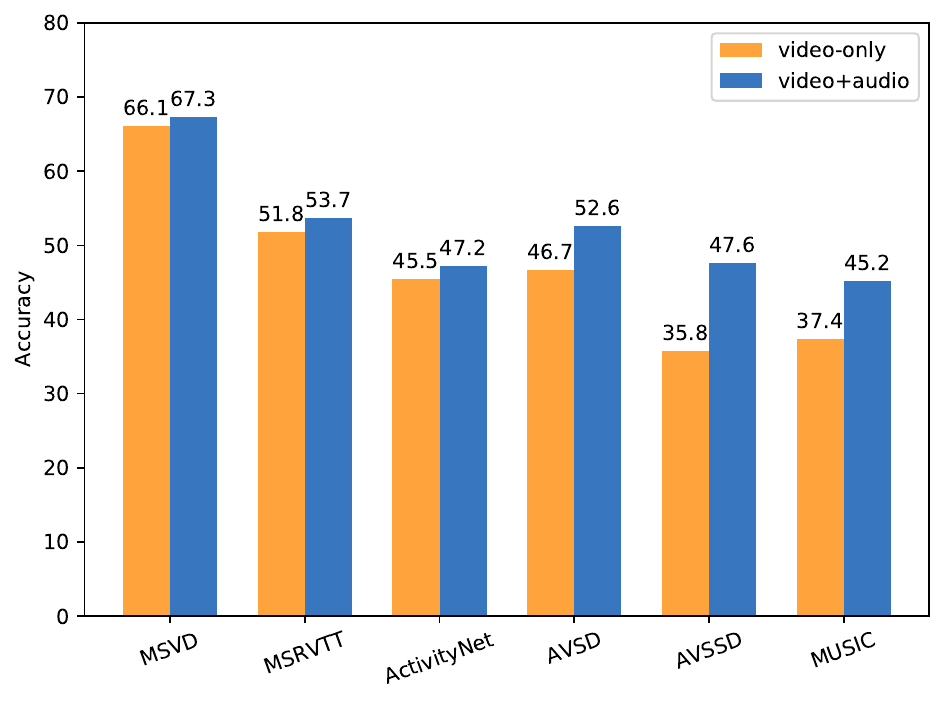}
    \caption{Ablation experiments on the integration of video modalities for video understanding. We report the comparison results on video QA and audio-visual QA between joint audio-visual modalities versus only the visual modality.}
    \label{fig:modality_intrgration}
\end{figure}

\subsection{Tuning Parameters}
The default configuration in our experiment is full tuning. This method requires a lot of training resources. Recent works explore freezing the model parameters and employing Low-Rank Adaptation (LoRA~\cite{lora}) to learn additional parameters for model fusion. We also compared LoRA with full tuning. The results in ~\cref{tab:tuning_params} show that LoRA can achieve comparable results with fewer GPU resources. However, there is still a gap compared to full tuning. We analyze that the limited number of parameters in LoRA may not be sufficient to store visual and audio information. Further research could investigate how to jointly train the visual encoder, audio encoder, and LLM with LoRA.

\begin{table}[h]
\centering
\scalebox{0.8}{
\begin{tabular}{l|ccc}
\toprule
\textbf{Tuning Params}  & \textbf{MSVD-QA}   & \textbf{MSRVTT-QA}  & \textbf{ActivityNet-QA} \\
\midrule
{LoRA}  & {62.8}  & {47.5} & {44.1}   \\
{Full Tuning}  & {67.3}  & {53.7}  & {47.2}  \\
\bottomrule
\end{tabular}}
\caption{Ablation experiments on the tuning parameters. We report results on video QA tasks between LoRA and full tuning.}
\label{tab:tuning_params}
\end{table}

\subsection{Compare on Multiple Dimensions}
The default evaluation focuses on a single dimension (accuracy). To fully evaluate the effectiveness of our method, we follow Video-ChatGPT and use GPT-4 to score (1$\sim$5) across multiple dimensions: Correctness of Information (Correct), Detail Orientation (Detail), Contextual Understanding (Context), Temporal Understanding (Temporal), and Consistency. The results in ~\cref{tab:diverse_dimensions} demonstrate that our method shows significant improvement over previous work, proving the efficacy of our method.

\begin{table}[h]
\centering
\scalebox{0.75}{
\begin{tabular}{l|ccccc}
\toprule
\textbf{Methods}  & \textbf{Correct}   & \textbf{Detail}  & \textbf{Context} & \textbf{Temporal} & \textbf{Consistency} \\
\midrule
{LLaMA-Adapter}  & {2.03}  & {2.32}  & {2.30} & {1.98} & {2.15} \\
{Video-LLaMA}  & {1.96} & {2.18} & {2.16}  & {1.82} & {1.79}  \\
{VideoChat}  & {2.23}  & {2.50}  & {2.53} & {1.94} & {2.24} \\
{Video-ChatGPT}  & {2.40}  & {2.52}  & {2.62} & {1.98} & {2.37} \\
{Valley}  & {2.43}  & {2.13}  & {2.86} & {2.04} & {2.45} \\
{Ours}  & {2.56}  & {2.47}  & {2.93} & {2.17} & {2.51} \\
\bottomrule
\end{tabular}}
\caption{GPT-based evaluation on multiple dimensions for video understanding. We follow Video-ChatGPT to report score results (1$\sim$5) on correct, detail, context, temporal, and consistency.}
\label{tab:diverse_dimensions}
\end{table}

\cleardoublepage
\thispagestyle{empty}

{
    \small
    \bibliographystyle{cvpr2024_conference}
    \bibliography{cvpr2024_conference}
}

 \label{fig:model-design}
\end{document}